\documentclass[letterpaper]{article}
\usepackage{aaai21}
\usepackage{times}
\usepackage{helvet}
\usepackage{courier}
\usepackage{multirow}

\usepackage[hyphens]{url}  
\usepackage{graphicx} 
\usepackage{natbib}  
\usepackage{caption} 
\usepackage{hyperref}
\usepackage{color}
\usepackage{graphicx} 
\usepackage{epsfig} 
\usepackage{subfigure}
\usepackage{makecell}

\usepackage{algorithm,algorithmic}
\usepackage{amsmath,amssymb,amsthm,amsfonts}
\usepackage{enumerate}
\usepackage{subfigure}

\usepackage{bm}
\usepackage{footnote}
\usepackage{url}

\title{Gradient Descent Averaging and Primal-dual Averaging \\ for Strongly Convex Optimization}

\newcommand*\samethanks[1][\value{footnote}]{\footnotemark[#1]}
\author{
    Wei Tao\textsuperscript{\rm 1,\rm 2},
	Wei Li\textsuperscript{\rm 2},
	Zhisong Pan\textsuperscript{\rm 2,}\thanks{Corresponding authors},
	Qing Tao\textsuperscript{\rm 3,}\samethanks\\
}
\affiliations{
     \textsuperscript{\rm 1}{Institute of Evaluation and Assessment Research, Academy of Military Science, Beijing 100091, China}\\
     \textsuperscript{\rm 2}{Command and Control Engineering College, Army Engineering University, Nanjing 210007, China}\\
     \textsuperscript{\rm 3}{Institute of Automation, Chinese Academy of Sciences, Beijing 100190, China}\\
    wtao\_plaust@163.com, liwei\_public@qq.com, hotpzs@hotmail.com, qing.tao@ia.ac.cn
}

\begin{document}
%

\maketitle

\begin{abstract}

Averaging scheme has attracted extensive attention in deep learning as well as traditional machine learning. It achieves theoretically optimal convergence and also improves the empirical model performance. However, there is still a lack of sufficient convergence analysis for strongly convex optimization. Typically, the convergence about the last iterate of gradient descent methods, which is referred to as individual convergence, fails to attain its optimality due to the existence of logarithmic factor. In order to remove this factor, we first develop \textit{gradient descent averaging} (GDA), which is a general projection-based dual averaging algorithm in the strongly convex setting. We further present \textit{primal-dual averaging} for strongly convex cases (SC-PDA), where primal and dual averaging schemes are simultaneously utilized. We prove that GDA yields the optimal convergence rate in terms of output averaging, while SC-PDA derives the optimal individual convergence. Several experiments on SVMs and deep learning models validate the correctness of theoretical analysis and effectiveness of algorithms.

\end{abstract}

\section{Introduction}

\noindent Averaging scheme has been widely adopted from different 
angles. It always helps to reduce variance and improve generalization of learning algorithms. In fact, there exist various averaging techniques, such as dual averaging (DA) \cite{nesterov2009primal}, weight averaging (WA) \cite{izmailov2018averaging}, output averaging (OA) \cite{nemirovsky1983problem, polyak1992acceleration}, primal averaging (PA) \cite{nesterov2015quasi,tao2020primal}, etc. 

DA was initially proposed by Nesterov \cite{nesterov2009primal}, which averages all past gradient information at each iteration. In comparison with gradient descent (GD) and mirror descent (MD) \cite{beck2003mirror}, it avoids new gradients to be considered with less weight than previous ones \cite{bach2017stochastic}. DA has been successfully extended to the stochastic composite scenario and it's well-suited for large-scale learning problems \cite{xiao2009dual,dekel2012optimal}. The superiority of regularized dual averaging (RDA) in efficiently promoting regularizer structure (e.g., sparsity) has been elaborated by Xiao and also earned test of time award at NeurIPS \cite{xiao2009dual}. 

Recently, averaging has also been frequently employed in training deep neural networks. WA averages weights of the networks based on training epochs \cite{izmailov2018averaging}. Since then, a series of contribution: SWALP \cite{yang2019swalp}, Fast-SWA \cite{athiwaratkun2018there}, SWA-Guassian \cite{maddox2019simple} have been successfully applied to a wide range of applications. Besides, exponential moving average (EMA), which has been used to exponentially decay the weights for previous iterate, can be regarded as a particular example of WA \cite{kingma2014adam, reddi2019convergence, ma2018quasi}.

OA is a classical way about how to output the final solution for iterative algorithms. Existing convergence analyses mostly center on it due to some superior theoretical guarantees \cite{bertsekas2003convex}. Actually, running algorithms for $t$ iterations, and returning the last iterate, is a very intuitive idea in practice. Therefore, there are still some gaps about individual output between theoretical analyses and practical implementations. Several works on stochastic gradient descent (SGD) develop different OA techniques to achieve the optimal convergence rate, especially for strongly convex optimization, such as suffix averaging \cite{rakhlin2011making}, non-uniform averaging \cite{lacoste2012simpler, harvey2019simple}, increasing weighted averaging \cite{guo2020revisiting}, etc. 

The optimal convergence for strongly convex problems has become a challenging problem after the well-known work \cite{hazan2006logarithmic}. This is because conventional SGD cannot attain the optimal convergence even when we take the uniform average of all past iterates. An open question early posed by \cite{shamir2012open} is that whether OA is needed at all to attain optimal convergence rate. Partially addressing this question, Shamir et al. \cite{shamir2013stochastic} showed that SGD with polynomial-decay averaging has an $O({\log t}/{\sqrt{t}})$ individual convergence rate in the general strongly convex cases and an $O({\log t}/{t})$ rate in strongly convex cases, respectively. Recent works \cite{harvey2019tight, jain2019making} provide the affirmative answer that the logarithmic term in the convergence bound is necessary for any plain SGD in both general convex and strongly convex cases. However, they leave us a new challenging problem. Can we achieve the optimal rate of $O(1/t)$ by slightly modifying any classical algorithms? From these observations, there are mainly two ways to achieve the optimal rate without the logarithmic factor. One is to modify the original steps of the algorithms. The other is to employ the averaging strategy.

PA is an interesting gradient operation step, in which the gradient evaluation is imposed on the weighted average of all past iterative solutions \cite{tao2020primal}. In fact, this averaging scheme was first used in PDA \cite{nesterov2015quasi}, which exploits simultaneously in both primal and dual space per-iteration, and succeeds in deriving the optimal individual rate for minimizing non-smooth general convex objectives. Later, \cite{tao2020primal} formally named it as PA, and they focus on projected subgradient (PSG) method. Its individual convergence rate doesn't suffer from the extra logarithmic factor. Overall, PDA is the closest solution to eliminate the $\log t$ factor about DA. Unfortunately, \cite{nesterov2015quasi} partially addressed the optimal convergence problem only in the general convex scenario. Optimal-RDA \cite{chen2012optimal} proposed earlier than PDA, requires two gradient operations per-iteration, which is different from conventional DA with only one operation. Similarly, \cite{cutkosky2019anytime} and \cite{joulani2020simpler} add an auxiliary PA scheme into their algorithms, which are able to achieve the optimal regret bound in the online setting. 


This paper is motivated by the breakthrough work of \cite{nesterov2015quasi}. Our original intention is to derive the optimal individual convergence of DA with minor changes in gradient operations. The main contributions can be summarized as follows:
\begin{itemize}

\item We present a general GDA algorithm, which includes the strongly convex algorithm in \cite{cutkosky2019anytime} as one special case of our method. Our GDA gains a deeper insight into the connection between DA and GD. Moreover, we prove that this algorithm no longer suffers from the logarithmic factor and attains the optimal convergence rate $O({1}/{t})$ coupled with OA. 

\item We incorporate PA into GDA, and develop a novel SC-PDA algorithm so as to achieve optimal individual convergence rate $O({1}/{t})$. Moreover, our convergence analysis of SC-PDA is obviously different from Nesterov's PDA. Thus, our work theoretically completes the task about individual convergence of DA under different convexity situations.

\end{itemize}

\section{Preliminaries and Notations}

Many convex optimization algorithms in machine learning can be formulated as a constrained black-box problem:
\begin{equation}\label{constrained minimization}
\min f(\mathbf{w}), \ s. t. \  \mathbf{w} \in \mathbf{Q}.
\end{equation}
where $\mathbf{Q}$ is a bounded convex domain, and $f\left(  \cdot  \right)$ is a convex function on $\mathbf{Q}$. Denote that $\mathbf{w}_{*}$ is an optimal solution. We use $\nabla f(\mathbf{w})$ to denote the (sub)gradient of $f$ at $\mathbf{w}$ and $\hat{\mathbf{g}}$ is an unbiased estimate of (sub)gradient of $f$ at $\mathbf{w}$.

Following \cite{shamir2013stochastic}, we first provide the definitions of strong convexity, individual convergence, and averaged convergence. 

{\bf Definition 1.} A function $f$ is called $\mu$-strongly convex with respect to the norm $\| \cdot \|$ if there is a constant $\mu > 0$ such that 
\begin{equation}\label{def.quatratic lower bound}
f(\mathbf{w}) \geq f(\mathbf{u})+\langle\nabla f(\mathbf{u}),\mathbf{w}-\mathbf{u}\rangle + \dfrac{\mu}{2} \|\mathbf{w}-\mathbf{u}\|^2,
\end{equation}
for all $\mathbf{u}, \mathbf{w}$.

Note that the strong convexity parameter $\mu$ is a measure of the curvature of $f$, $\left< \cdot,\cdot \right>$ stands for the Euclidean inner product, and the quadratic lower bound in (\ref{def.quatratic lower bound}) can be also satisfied with $\mu = 0$ for a convex function.

Generally, the convergence about the last iterate is often referred to as \textit{individual convergence} for simplicity \cite{tao2020primal}.

{\bf Definition 2.} Given a convex function $f$, let $\{\mathbf{w}_t\}_{t>0}$ be generated by optimization algorithms, the individual convergence is defined as
\begin{equation}\label{def.individual convergence}
f(\mathbf{w}_{t}) - f(\mathbf{w}_{*}) \leq \epsilon(t).
\end{equation}

{\bf Definition 3.} Given a convex function $f$ with uniform averaged output $\mathbf{\bar w}_{t} = \frac{1}{t} \sum_{i=1}^{t} \mathbf{w}_{i} $, let $\mathbf{w}_t$ be generated by optimization algorithms, we can define averaged convergence as
\begin{equation}\label{def.averaged convergence}
f(\mathbf{\bar w}_{t}) - f(\mathbf{w}_{*}) \leq \epsilon(t).
\end{equation}
The convergence bound $\epsilon$ is related to $t$. In particular, the optimal bound is $O({1}/{\sqrt{t}})$ in the non-smooth convex cases and $O({1}/{t})$ in the strongly convex cases, respectively \cite{nesterov27method, nemirovsky1983problem}.

\section{Related Work}

In this section, we briefly review some related algorithms and their convergence rates. PSG is one of the most fundamental algorithms for solving (\ref{constrained minimization}), and the iteration of which is,
\begin{equation}\label{Eq.psg}
\mathbf{w}_{t+1} = P[\mathbf{w}_{t} - a_t  \nabla f(\mathbf{w}_t)],
\end{equation}
where $P$ is the projection operator on $ \mathbf{Q} $, $a_t>0$ is the step-size parameter. More generally, mirror descent (MD) is a direct extension of the PSG by using a mirror map, and it iterates as follows,
\begin{equation}\label{MD}
{\mathbf{w}}_{t+1}=\arg\min_{\mathbf{w} \in \mathbf{Q}} \{a_t\langle\nabla f(\mathbf{w}_t),\mathbf{w}\rangle+B(\mathbf{w},\mathbf{w}_t)\},
\end{equation}
where $t>0$, $B$ is the Bregman divergence. MD recovers PSG by taking $\frac{1}{2}\|\mathbf{w}-\mathbf{w}_{t}\|^2$. 

Based on MD, DA is also a powerful first-order gradient algorithm \cite{nesterov2009primal}. The standard DA updates the solution according to
\begin{equation}\label{Eq.DA}
{\mathbf{w}}_{t+1}=\arg\min_{\mathbf{w} \in \mathbf{Q}} \{\sum_{k=0}^{t}\langle a_{k}\nabla f(\mathbf{w}_k),\mathbf{w}\rangle+\gamma_t d(\mathbf{w})\},
\end{equation}
where $a_t$ is the weight parameter, $\gamma_t>0$ is the stepsize,  $d(\cdot)$ is a strongly convex function such that
$$ d(\mathbf{w}) \geq d(\mathbf{u})+\langle\nabla d(\mathbf{u}),\mathbf{w}-\mathbf{u}\rangle +\frac{1}{2}\|\mathbf{w-u}\|^2, \forall \mathbf{u}, \mathbf{w} \in \mathbb{R}^{n}.$$
It has been shown in \cite{nesterov2015quasi} that,
\begin{equation}\label{Eq.da bound}
\begin{aligned}
\frac{1}{A_t} & \sum_{k=0}^{t}a_k f(\mathbf{w}_k)-f(\mathbf{w}_*)
\\ & \leq  \frac{1}{A_t}\big[\gamma_t d(\mathbf{w}_*) + \sum_{k=0}^{t}\frac{a_{k}^{2}}{2\gamma_k}\|\nabla f(\mathbf{w}_k)\|^2 \big].
\end{aligned}
\end{equation}
where $A_t = \sum_{k=0}^{t}a_k$. 

RDA is a proximal variant of DA algorithm that dramatically captures the geometry structure of the regularizer for stochastic composite learning problems. The key iteration is as follows,
\begin{equation}\label{Eq.RDA}
\mathbf{w}_{t+1} = \arg\min_{\mathbf{w} \in \mathbf{Q}} \{\frac{1}{t}\langle \sum_{k=0}^{t} \nabla f(\mathbf{w}_k), \mathbf{w}\rangle+ r(\mathbf{w}) +\frac{\beta_t}{t}d(\mathbf{w})\},
\end{equation}
where $r(\mathbf{w})$ is the regularization function, $d(\mathbf{w})$ is a strongly convex regularization term, $\{ \beta_t \}_{t>0} $ is a non-negative and non-decreasing sequence. RDA (\ref{Eq.RDA}) accumulates the weight of a sparse regularizer such as $l_1$-norm to produce more sparse solutions. By averaging all past gradient instead of using current gradient information, DA always exhibits  more stable convergence behavior. Besides, it uses a global proximal function $d(\mathbf{w})$ as opposed to local Bregman divergence $B(\mathbf{w},\mathbf{w}_t)$ in (\ref{MD}).

In order to obtain the optimal individual convergence rate, \cite{nesterov2015quasi} developed PDA where the primal and dual averaging schemes are simultaneously employed. It can be viewed as an alternative answer to the Shamir's open problem by slightly modifying DA. The key steps can be described as follows,
\begin{subequations}\label{Eq.qmDA}
\begin{align}
& \mathbf{w}_t^+ = \arg\min_{\mathbf{w} \in \mathbf{Q}} \{\sum_{k=0}^{t}\langle a_{k}\nabla f(\mathbf{w}_k),\mathbf{w}\rangle+\gamma_t d(\mathbf{w})\},\\
& \mathbf{w}_{t+1} = \frac{A_t}{A_{t+1}} \mathbf{w}_t + \frac{a_{t+1}}{A_{t+1}}\mathbf{w}_t^+.
\end{align}
\end{subequations}
Obviously, the only difference between standard DA (\ref{Eq.DA}) and PDA (\ref{Eq.qmDA}) lies in the additional weighted averaging step (\ref{Eq.qmDA}b), also called PA \cite{tao2020primal, taylor2019stochastic}. Only considering the general convex case, it was proved to derived the optimal individual convergence,
\begin{equation}\label{Eq.DA individual}
f(\mathbf{w}_t)-f(\mathbf{w}_*)  \leq  \frac{1}{A_t}[\gamma_t d(\mathbf{w}_*) + \sum_{k=0}^{t}\frac{a_{k}^{2}}{2\gamma_k}\|\nabla f(\mathbf{w}_k)\|^2].
\end{equation}

Based on the PDA, \cite{tao2020primal} presented PA-PSG for both general convex (\ref{Eq.PA-PSG-general}) and strongly convex scenarios (\ref{Eq.PA-PSG-strong}), such that
\begin{subequations}\label{Eq.PA-PSG-general}
\begin{align}
& \mathbf{w}_t^+ = P[\mathbf{w}_{t-1}^+ - a_t  \nabla f(\mathbf{w}_t)],\\
& \mathbf{w}_{t+1} = \frac{A_t}{A_{t+1}} \mathbf{w}_t + \frac{a_{t+1}}{A_{t+1}}\mathbf{w}_t^+,
\end{align}
\end{subequations}
and 
\begin{subequations}\label{Eq.PA-PSG-strong}
\begin{align}
& \mathbf{w}_t^+ = P[\delta_t \mathbf{w}_{t-1}^+ - {a_t}{\delta_t} (\nabla f(\mathbf{w}_t)- \mu \mathbf{w}_t) ],\\
& \mathbf{w}_{t+1} = \frac{A_t}{A_{t+1}} \mathbf{w}_t + \frac{a_{t+1}}{A_{t+1}}\mathbf{w}_t^+,
\end{align}
\end{subequations}
where $\delta_t = 1/(1+a_t\mu)$.

Note that the gradient operations are imposed on $f(\mathbf{w}_t)$, and now $\mathbf{w}_t$ becomes a weighted average of all past iterative primal sequences. PA-PSG also achieves the following convergence rate, 
\begin{equation}\label{da individual bound}
f(\mathbf{w}_t)-f(\mathbf{w}_*)  \leq  \frac{1}{A_t}[B(\mathbf{w}_0, \mathbf{w}_*) + \sum_{k=0}^{t}\frac{a_{k}^{2}}{2}\|\nabla f(\mathbf{w}_k)\|^2].
\end{equation}
It reveals that PA-PSG in the convex setting converges to the optimum at $O({1}/{\sqrt{t}})$ \cite{tao2020primal}. Besides, \cite{defazio2020factorial} also established connections between PA and momentum methods. 

Online-to-batch conversion is a standard way to obtain convergence guarantees from online learning algorithms to stochastic convex optimization \cite{shalev2007pegasos,hazan2011beyond}. Recently, \cite{cutkosky2019anytime} developed anytime online learning algorithms whose last iterate converge to the optimum in the stochastic cases. \cite{cutkosky2019anytime} focused on DA, and the strongly convex algorithm can be described as follows,
\begin{equation}\label{Eq.anytime}
\mathbf{w}_{t+1} = \arg\min_{\mathbf{w} \in \mathbf{Q}} \{\langle \sum_{k=0}^{t} \nabla f(\mathbf{w}_k), \mathbf{w}\rangle + \frac{\mu}{2} \sum_{k=0}^{t}  \| \mathbf{w}-\mathbf{w}_k \|^2 \}.
\end{equation}
It has also been proved that
\begin{equation}\label{Eq.anytime bound}
\mathbb{E}[f(\mathbf{w}_t)-f(\mathbf{w}) ]\leq O\left(\frac{\log t}{2\mu t}\right).
\end{equation}
Obviously, this convergence bound (\ref{Eq.anytime bound}) for stochastic optimization is suboptimal.

Based on the above observations, we can find that the optimal individual convergence rate of DA for strongly convex optimization problem is still missing, which is one of our main motivations in this paper.

\section{Proposed GDA and SC-PDA Algorithms}

In this section, we focus on the case where $f$ satisfies $\mu$-strongly convex, and present two modified projection-based DA algorithms for solving (\ref{constrained minimization}), then we will analyse their performance in terms of optimal convergence rate. The proofs of theoretical results in this section are exhibited in the supplementary material.

\subsection{GDA and Averaged Convergence}

Following the work of RDA for strongly convex functions \cite{xiao2009dual}, we replace the term $\gamma_t d(\mathbf{w})$ in standard DA (\ref{Eq.DA}) with another $\mu$-strongly convex term $\frac{\mu}{2}\sum_{k=0}^{t}\gamma_k \|\mathbf{w}-\mathbf{w}_k \|^2$, then we have
\begin{equation}\label{Eq.gda}
\begin{aligned}
\mathbf{w}_{t+1}=\arg\min_{\mathbf{w} \in \mathbf{Q}} \{\sum_{k=0}^{t} & \langle a_{k}\nabla f(\mathbf{w}_k), \mathbf{w}\rangle
\\ & + \dfrac{\mu}{2}\sum_{k=0}^{t}\gamma_k \|\mathbf{w}-\mathbf{w}_k \|^2 \}.
\end{aligned}
\end{equation}
We can find that the strongly convex algorithm (\ref{Eq.anytime}) is one special case of our GDA algorithm  (\ref{Eq.gda}) when we let $a_k = 1$ and $\gamma_k =1$.

The detailed steps of GDA algorithm are shown in Algorithm \ref{Alg.GDA}. According to the following lemmas, we can also get the projection version of DA. We first provide the property of projection operator,

{\bf Lemma 1.} \cite{bertsekas2003convex} For $\mathbf{w} \in \mathbb{R}^{n}, \mathbf{w}_0 \in \mathbf{Q}$,
$$
\langle \mathbf{w}-\mathbf{w}_0, \mathbf{u}-\mathbf{w}_0\rangle \leq 0,
$$
for all $\mathbf{u} \in \mathbf{Q}$ if $\mathbf{w}_0 = P(\mathbf{w})$.

In the following lemma, we can get the equivalent form between the projection method and the proximal method.

{\bf Lemma 2.} For $\mathbf{w} \in \mathbb{R}^{n}$, standard DA algorithm (\ref{Eq.DA}) is equivalent to 
\begin{equation}\label{Eq.da-pro}
\mathbf{w}_{t+1} = P \big[ - \frac{\sum_{k=0}^{t}a_k \nabla f(\mathbf{w}_k)}{\gamma_t}\big],
\end{equation}
where $P$ is the projection operator on $\mathbf{Q}$.

Based on Lemma 1, we can get the following projection-based DA for strongly convex objectives, which is also equivalent to (\ref{Eq.gda}), i.e.,
\begin{equation}\label{Eq.gda-pro}
\mathbf{w}_{t+1} = P\big\{ \frac{1}{\Gamma_t} \sum_{k=0}^{t} \gamma_k[ \mathbf{w}_{k}-  \frac{a_k}{\mu\gamma_k}\nabla f(\mathbf{w}_{k})]  \big\},
\end{equation}
where $t > 0$, $\Gamma_t = \sum_{k=0}^{t} \gamma_k$. The stepsize is ${a_k}/(\mu\gamma_k)$. Notably, this algorithm (\ref{Eq.gda-pro}) can be viewed as a weighted averaging GD, which we name GDA in the paper. In contrast to PIWA \cite{guo2020revisiting}, our original intention is to slightly modify DA, and then we build the connection between DA and GD. Besides, PA is different from the weighted averaging scheme employed in PIWA. 

\begin{algorithm}
	\renewcommand{\algorithmicrequire}{\textbf{Input:}}
	\renewcommand{\algorithmicensure}{\textbf{Averaged Output:}}
	\caption{GDA}
	\label{Alg.GDA}
	\begin{algorithmic}[1]
		\REQUIRE strongly convex parameter $\mu$, non-negative stepsize sequence $a_t$ and $\gamma_t$.
		\STATE Initialize $\mathbf{w}_0 = \mathbf{0}$, $a_0 = 0, \gamma_0=\Gamma_0=0$.
		\REPEAT
		\STATE Calculate $\Gamma_{t+1} = \Gamma_t + \gamma_{t+1}$,
		\STATE Compute subgradient of 
		$\nabla f(\mathbf{w}_t)$,
		\STATE Update (sub)gradient descent step:\\ $\mathbf{w}_{t1} =  \mathbf{w}_{t}- \frac{a_t}{\gamma_t \mu} \nabla f(\mathbf{w}_t)$,
		\STATE Update weighted averaging step:\\ 
		$\mathbf{w}_{t2} =  \frac{1}{\Gamma_t} \sum_{k=0}^{t} \gamma_k (\mathbf{w}_{k1})$,
       \STATE Update projection step:\\
       $\mathbf{w}_{t+1} = P(\mathbf{w}_{t2})$,
		\UNTIL{convergence} 
		\ENSURE $\bar{\mathbf{w}}_{t+1}$
	\end{algorithmic}  
\end{algorithm}

To conduct convergence analysis, we also need the following assumptions about the gradient oracle.

{\bf Assumption 1.} The (sub)gradient oracle is $M$-bounded with $M$ \textgreater 0, i.e.,
\begin{equation}\label{Assu.gradinet bounded}
\|\nabla f(\mathbf{w})\| \leq M.
\end{equation}

{\bf Assumption 2.} Let $\hat{\mathbf{g}}_t$ be an unbiased estimate of subgradient of $f$ at $\mathbf{w}$. For any $\mathbf{w}$, 
\begin{equation}\label{Assu.gradinet}
\| \hat{\mathbf{g}}_t -\nabla f(\mathbf{w})\| \leq \sigma^2.
\end{equation}

Then, we can get the following theorem.

{\bf Theorem 1.} Let $\{{\mathbf{w}}_{t}\}_{t=1}^{\infty}$ be generated by algorithm (\ref{Eq.gda-pro}). For any $\mathbf{w} \in \mathbf{Q}$, we have
$$
\begin{aligned}
& \frac{1}{A_t} \sum_{k=0}^{t}a_k f(\mathbf{w}_k)-f(\mathbf{w}_*) \\
& \leq  \frac{1}{2A_t}\sum_{k=0}^{t}\big[\frac{a_{k}^{2}}{\mu\Gamma_{k}}\|\nabla f(\mathbf{w}_k)\|^2 + \mu (\gamma_k-a_k) \| \mathbf{w}_k-\mathbf{w}_*\|^2 \big]
\end{aligned}
$$
where $A_t = \sum_{k=0}^{t}a_k$, $\Gamma_t = \sum_{k=0}^{t}\gamma_k$. 

{\bf Corollary 1.} Let ${\mathbf{w}}_0 \in \mathbf{Q} $ be the initial point and $\{{\mathbf{w}}_{t}\}_{t=1}^{\infty}$ be generated by GDA (\ref{Eq.gda-pro}). There exists a positive number $M_0 > 0$ such that
$$
\| \mathbf{w}_t - \mathbf{w}_*\| \leq M_0.
$$
Specifically, let $a_t=\gamma_t=t$. For any $\mathbf{w} \in \mathbf{Q}$, it holds that
$$f(\bar{\mathbf{w}}_t)-f(\mathbf{w}_*)\leq O\left(\frac{1}{t}\right).$$

{\bf Remark 1.} For problem (\ref{constrained minimization}), Theorem 1 indicates that GDA achieves the  optimal rate $O({1}/{t})$ as opposed to the suboptimal rate with the multiplicative $\log t$ factor in (\ref{Eq.anytime bound}). With suitably chosen stepsize and strongly convex parameter $\mu$ and $\gamma_t$, we get the optimal averaged convergence of GDA. However, it remains unclear the optimal individual convergence rate of DA for strongly convex optimization problems. 

\subsection{SC-PDA and Individual Convergence}

Based on GDA, we discuss the detail steps of our second approach (Algorithm 2) below and illustrate how to achieve the optimal individual convergence rate without the extra logarithmic factor.

Motivated by their work \cite{nesterov2015quasi, tao2020primal}, we incorporate PA into DA for strongly convex objective functions. The key iterations are given by
\begin{subequations}\label{Eq.doa}
\begin{align}
& {\mathbf{w}}_{t}^+=\arg\min_{\mathbf{w} \in \mathbf{Q}} \{\sum_{k=0}^{t}\langle a_{k}\nabla f(\mathbf{w}_{k}),\mathbf{w}\rangle + \frac{\mu}{2}\sum_{k=0}^{t}\gamma_{k} \|\mathbf{w}-\mathbf{w}_{k} \|^2 \}, \\
& \mathbf{w}_{t+1} = \frac{A_t}{A_{t+1}} \mathbf{w}_t + \frac{a_{t+1}}{A_{t+1}}\mathbf{w}_t^+,
\end{align}
\end{subequations}
where $t>0$, and $A_t = \sum_{k=0}^{t}a_k$. Also, there is the equivalent projection-type algorithm,
\begin{subequations}\label{Eq.pda}
\begin{align}
& \mathbf{w}_t^+ = P\big\{ \frac{1}{\Gamma_t} \sum_{k=0}^{t} \gamma_k[ \mathbf{w}_{k}-  \frac{a_k}{\mu\gamma_k}\nabla f(\mathbf{w}_{k})]  \big\},\\
& \mathbf{w}_{t+1} = \frac{A_t}{A_{t+1}} \mathbf{w}_t + \frac{a_{t+1}}{A_{t+1}}\mathbf{w}_t^+.
\end{align}
\end{subequations}
We can find that (\ref{Eq.pda}a) and (\ref{Eq.pda}b) are both averaging steps in the strongly convex setting, which is called SC-PDA in this paper.

\begin{algorithm}
	\renewcommand{\algorithmicrequire}{\textbf{Input:}}
	\renewcommand{\algorithmicensure}{\textbf{Individual Output:}}
	\caption{SC-PDA}
	\label{Alg.pda}
	\begin{algorithmic}[1]
		\REQUIRE strongly convex parameter $\mu$. stepsize parameter $a_t$ and $\gamma_t$,
		\STATE Initialize $\mathbf{w}_0 = \mathbf{w}_0^+ = \mathbf{0}$, $a_0 = A_0 = 0$, $\gamma_0 = \Gamma_0 = 0$.
		\REPEAT
		\STATE Calculate $A_{t+1} = A_{t} + a_{t+1}$, $\Gamma_{t+1} = \Gamma_t + \gamma_{t+1}$,
		\STATE Compute subgradient of 
		$\nabla f(\mathbf{w}_t)$
        \STATE Update (sub)gradient descent step:\\ $\mathbf{w}_{t1}^+ =  \mathbf{w}_{t}- \frac{a_k}{\mu \gamma_k} \nabla f(\mathbf{w}_t)$,
		\STATE Update weighted averaging step:\\
		$\mathbf{w}_{t2}^+ =  \frac{1}{\Gamma_t} \sum_{k=0}^{t} \gamma_k (\mathbf{w}_{k1}^+)$,
        \STATE Update projection step:\\
        $\mathbf{w}_{t}^+ = P(\mathbf{w}_{t2}^+)$,
        \STATE Update primal averaging step:\\
		$\mathbf{w}_{t+1} = \frac{A_{t}}{A_{t+1}} \mathbf{w}_t + \frac{a_{t+1}}{A_{t+1}}\mathbf{w}_{t}^+$.
		\UNTIL{convergence} 
		\ENSURE $\mathbf{w}_{t+1} $
	\end{algorithmic}  
\end{algorithm}

The individual convergence rates are based on the following lemma and theories. Lemma 3 bridges the connection between variable $\mathbf{w}$ and its corresponding objective function $f(\mathbf{w})$. 




{\bf Lemma 3.} Assume $f(\mathbf{w})$ is a strongly convex function. Then let $\{{\mathbf{w}}_{t}\}_{t=1}^{\infty}$ be generated by algorithm (\ref{Eq.pda}). For any $\mathbf{w}  \in \mathbf{Q}$, we have 	
$$
\begin{aligned}
 &  a_t \langle \nabla f(\mathbf{w}_t),\mathbf{w}_t - \mathbf{w}_t^+ \rangle
\\ & \leq A_{t-1} [f(\mathbf{w}_{t-1}) - f(\mathbf{w}_{t})] + a_t \langle \nabla f(\mathbf{w}_t),\mathbf{w}_{t-1}^+ - \mathbf{w}_{t}^+\rangle.
\end{aligned}
$$

{\bf Remark 2.} Lemma 3 bridges the connection between $\mathbf{w}$ and $f(\mathbf{w})$. In other words, Lemma 3 is the most important point in deriving individual convergence rate. Based on Lemma 1-3, we can obtain the following theorem.

{\bf Theorem 2.} Let $\{{\mathbf{w}}_{t}\}_{t=1}^{\infty}$ be generated by algorithm (\ref{Eq.pda}). For any $\mathbf{w} \in \mathbf{Q}$, we have
$$
\begin{aligned}
& f(\mathbf{w}_t)-f(\mathbf{w}_*) \\ 
& \leq  \frac{1}{2A_t}\sum_{k=0}^{t}[\frac{a_{k}^{2}}{\mu\Gamma_{k}}\|\nabla f(\mathbf{w}_k)\|^2 + \mu (\gamma_k-a_k) \| \mathbf{w}_k-\mathbf{w}_*\|^2]
\end{aligned}
$$

Further, we also can get the following Corollaries.

{\bf Corollary 2.} Let $\{{\mathbf{w}}_{t}\}_{t=1}^{\infty}$ be generated by our proposed algorithm (\ref{Eq.pda}). According to the Assumption 1 about (sub)gradient bound.

(1) Let $a_t=\gamma_t= t$. For any $\mathbf{w} \in \mathbf{Q}$, it holds that
\begin{equation}\label{Eq.PDA_individual1}
f(\mathbf{w}_t)-f(\mathbf{w}_*)\leq O\left(\frac{1}{t}\right). 
\end{equation}

(2) Let $a_t=\gamma_t=1$. For any $\mathbf{w} \in \mathbf{Q}$, we can get
\begin{equation}\label{Eq.PDA_individual2}
f(\mathbf{w}_t)-f(\mathbf{w}_*)\leq O\left(\frac{\log t}{t}\right).
\end{equation}

{\bf Remark 3.} Obviously, (\ref{Eq.PDA_individual1}) in Corollary 2 indicates that the optimal rate of individual convergence for strongly convex problems can be achieved by our proposed SC-PDA (\ref{Eq.pda}). With suitably chosen the stepsize parameters,  (\ref{Eq.PDA_individual2}) is same as the bound (\ref{Eq.anytime bound}). It should be noticed that the convergence analysis in our paper is quietly different from that in \cite{cutkosky2019anytime}.

It has been indicated in \cite{harvey2019tight} and \cite{jain2019making}, the suboptimal $O({\log t}/{t})$ individual convergence for standard GD is tight under strong convexity condition. As we have established the connection between these two first-order gradient algorithms, DA also exhibits the same individual convergence behavior. (\ref{Eq.PDA_individual1}) exposes that if we choose the suitable parameters, GDA is able to remove the extra logarithmic factor and accelerate the suboptimal convergence rate of DA.

{\bf Corollary 3.} According to (\ref{Eq.PDA_individual1}), we can easily attain the rate of averaged convergence, such that
\begin{equation}
f(\bar{\mathbf{w}}_t)-f(\mathbf{w}_*) \leq O\left(\frac{\log t}{t}\right).
\end{equation}

{\bf Remark 4.} In contrast to the convergence bound in Corollary 1, we can only obtain the averaged convergence rate with a $\log t$ factor through the optimal individual convergence rate. In other words, averaged convergence can't easily transform to the individual one. Thus, there are no transition relations between these two convergence rate for strongly convex optimization.

\section{Extension to the Stochastic Setting}

In this section, we consider the binary SVM problems for simplicity, and transform our deterministic approaches to stochastic versions. 

Due to the data explosion in recent years, deterministic optimization methods that need to evaluate a large number of full gradients are not suitable for solving very large-scale optimization problems. Stochastic optimization methods can alleviate this limitation by sampling one (or a small set of) examples and computing a stochastic (sub)gradient at each iteration based on the sampled examples. Therefore, we extend our method to the stochastic setting. 

Let the training set $S =\{( \mathbf{x}_1 , y_1), (\mathbf{x}_2 , y_2),...,(\mathbf{x}_n , y_n) \}$, where $\mathbf{x}_i $ is \textit{i.i.d} (independently identically distribution). $y_i \in Y = \{-1,1\}$ is the label. $(\mathbf{x}_i , y_i) \}$ is uniformly at random chosen from $S$. For problem (\ref{constrained minimization}) and stochastic algorithms, let $f(\mathbf{w}) = \sum_{i=1}^{n}f_i(\mathbf{w})$, $f_i(\mathbf{w}) = \max\{0, 1-y_i\langle \mathbf{w},\mathbf{x}_{i}\rangle\}$ is the non-smooth loss function of $(\mathbf{x}_i , y_i)$, $\mathbf{Q}$ is the closed convex set.

The key steps of the stochastic SC-PDA method are,
\begin{subequations}\label{stochastic PDA}
\begin{align}
& \mathbf{w}_t^+ = P\big\{ \frac{1}{\Gamma_t} \sum_{k=0}^{t} \gamma_{k} [ \mathbf{w}_{k}- \frac{a_k}{\mu \gamma_k}\hat{\mathbf{g}}_k] \big\},\\
& \mathbf{w}_{t+1} = \frac{A_t}{A_{t+1}} \mathbf{w}_t + \frac{a_{t+1}}{A_{t+1}}\mathbf{w}_t^+,
\end{align}
\end{subequations}
where $\hat{\mathbf{g}}_t$ is an unbiased estimate of subgradient of $f$ at $\mathbf{w}_t$, $\mu$ is the strongly convex parameter.

In the following, we will analyse the convergence rate in expectation of the stochastic SC-PDA method (\ref{stochastic PDA}). When demonstrating the convergence rate of stochastic optimization algorithms from their deterministic settings, one way is to replace the real gradient with its unbiased estimation, which is carefully described in \cite{rakhlin2011making}\footnote{We follow the proof of Lemma 1.}. Specifically, after adding a real gradient, we find that the term, developed by the gap between the real gradient and the stochastic gradient, is not affect the original convergence properties for non-smooth optimization problems. Therefore, it is easy to derive Theorem 3.

{\bf Theorem 3.} $f$ is $\mu$-strongly convex about the Bregman divergence $B$. Let ${\mathbf{w}}_{0} \in \mathbb{R}^{n} $ be the initial point and $\{{\mathbf{w}}_{t}\}_{t=1}^{\infty}$ be generated by the stochastic SC-PDA (\ref{stochastic PDA}). Lemma 3 becomes
$$
\begin{aligned}
 &  a_t \langle \hat{\mathbf{g}}_t,\mathbf{w}_t - \mathbf{w}_t^+ \rangle
\\ & \leq A_{t-1} [f(\mathbf{w}_{t-1}) - f(\mathbf{w}_{t})] + a_t \langle \hat{\mathbf{g}}_t,\mathbf{w}_{t-1}^+ - \mathbf{w}_{t}^+\rangle.
\end{aligned}
$$
Further, we have
$$
\begin{aligned}
& \mathbb{E}[f(\mathbf{w}_t)-f(\mathbf{w}_*)] \\ 
& \leq  \frac{1}{2A_t}\sum_{k=0}^{t}\big[\frac{a_{k}^{2}}{\mu\Gamma_{k}}\|\hat{\mathbf{g}}_k\|^2 + \mu (\gamma_k-a_k) \| \mathbf{w}_k-\mathbf{w}_*\|^2 \big].
\end{aligned}
$$

Then we have the following corollary, that is

{\bf Corollary 4.} Let $\{{\mathbf{w}}_{t}\}_{t=1}^{\infty}$ be generated by our stochastic algorithm (\ref{stochastic PDA}). Let $a_t=\gamma_t=t$, it holds that
\begin{equation}
\mathbb{E}[f(\mathbf{w}_t)-f(\mathbf{w}) ]\leq O\left(\frac{1}{t}\right).
\end{equation}

{\bf Remark 5.} According to Theorem 3, the optimal linear rate of individual convergence in the non-smooth setting is achieved by the stochastic SC-PDA (\ref{stochastic PDA}).

Based on the convergence analysis for SC-PDA, the optimal individual convergence rate for strongly convex problems is achieved. Therefore, we derive the conclusion that the OA is actually unnecessary for DA whereas we should modified the key steps of the algorithm and incorporate averaging schemes into it. As illustrated in \cite{jain2019making, harvey2019tight}, Shamir's open problem has been solved to some extent. In contrast to their results, our contribution center on a more complex first-order algorithm DA for non-smooth strongly convex optimization without priori knowledge of the performed number of iterations $T$.

\section{Experiments}
In this section, we conduct several experiments to verify our theoretical claims and demonstrate the performance of our proposed algorithms in training deep networks. 

\subsection{Optimizing Strongly Convex Functions}

In the first experiment, we consider classical binary strongly convex SVM problems.
\begin{equation}\label{SVM}
\min_{\mathbf{w}}  \frac{\mu}{2} \|\mathbf{w}\|^2+ \sum_{i=0}^{n} f_i (\mathbf{w}),
\end{equation}
where $f_{i}(\mathbf{w}) = \max\{0, 1-y_i\langle \mathbf{w},\mathbf{x}_{i}\rangle\}$.

We choose four benchmark datasets: a9a, w8a, covtype, ijcnn1 with different scale and dimension, which are publicly available at \textit{LibSVM}\footnote{\url{http://www.csie.ntu.edu.tw/~cjlin/libsvmtools/datasets/}} website. We choose the best stepsize parameter via commonly used grid search technique. For fair comparison, we independently repeated the experiments five times, and averaged the results. In stochastic learning, at the $t$-the iterations,
\begin{equation}
\hat{\mathbf{g}}_t=\nabla f_{t}(\mathbf{w}_t),
\end{equation}
where the sample $(\mathbf{x}_{t},y_{t})$ is uniformly at random chosen from the training set. 

Here, we compare our GDA (\ref{Eq.gda-pro}) and SC-PDA (\ref{Eq.pda}) with three state-of-the-art stochastic approaches for strongly convex functions to validate our theoretical analysis.
\begin{itemize}
\item GDA: the proposed method doesn't suffer from the extra $\log t$ factor in the bound, and yield the optimal convergence $O(1/t)$ in terms of OA.
\item SC-PDA: GDA coupled with PA attains the optimal individual convergence rate $O(1/t)$.
\item Pegasos \cite{shalev2007pegasos}: PSG outputs the individual iterate. Here, the decreasing stepsize is $1/(\mu t)$. As illustrated in \cite{harvey2019tight}, standard pegasos can obtain at most a rate $O(\log t /t)$ with high probability.
\item PA-PSG \cite{tao2020primal}: PSG coupled with PA achieves the individual convergence rate $O(1/t)$.
\item SC-RDA \cite{xiao2009dual}: RDA combined with OA for strongly convex optimization obtains the optimal rate of convergence $O(1/t)$.
\end{itemize}

\begin{figure*}[ht]
    \centering
    \subfigure[]{
    \includegraphics[width=1.6in]{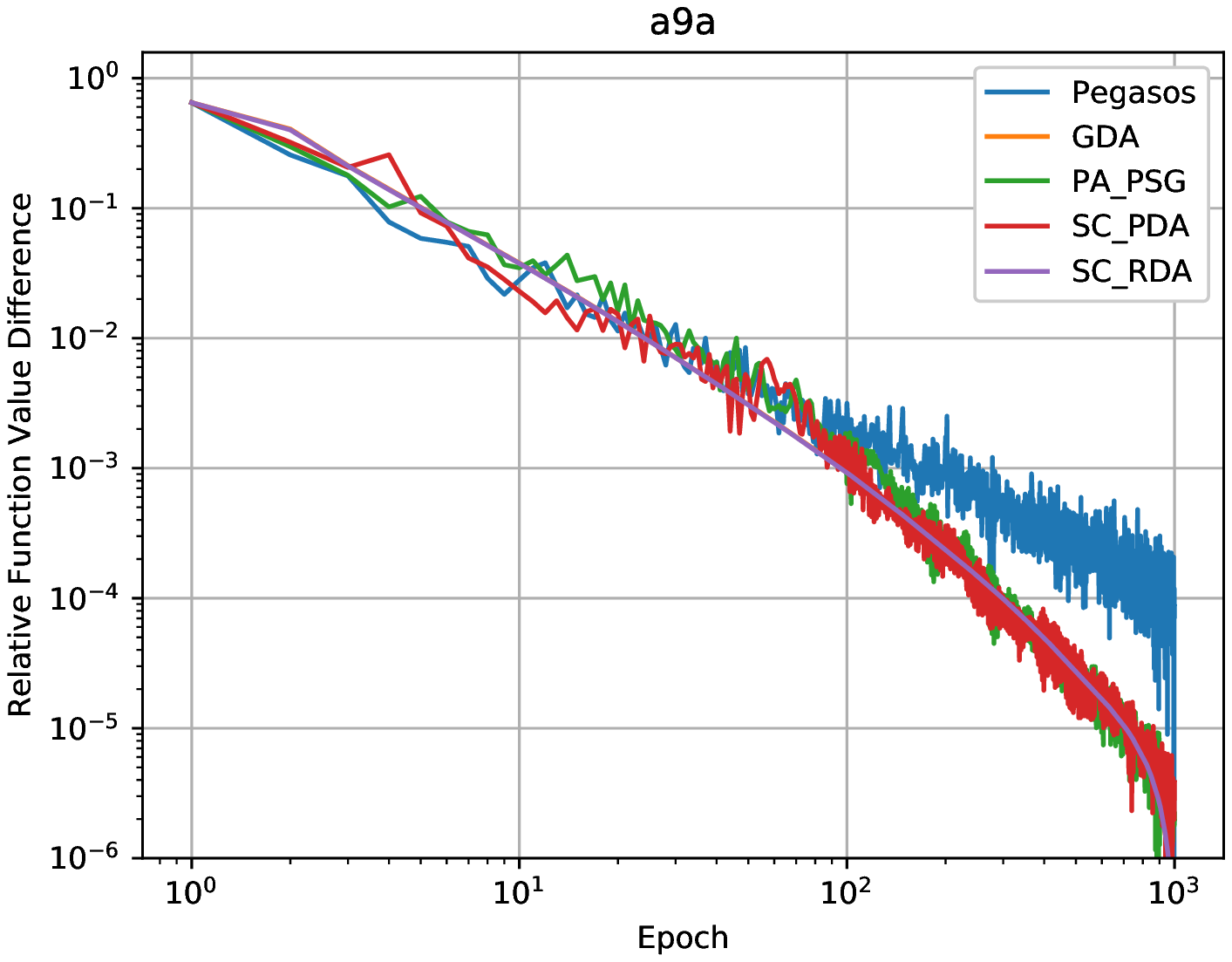}}
    \subfigure[]{
    \includegraphics[width=1.6in]{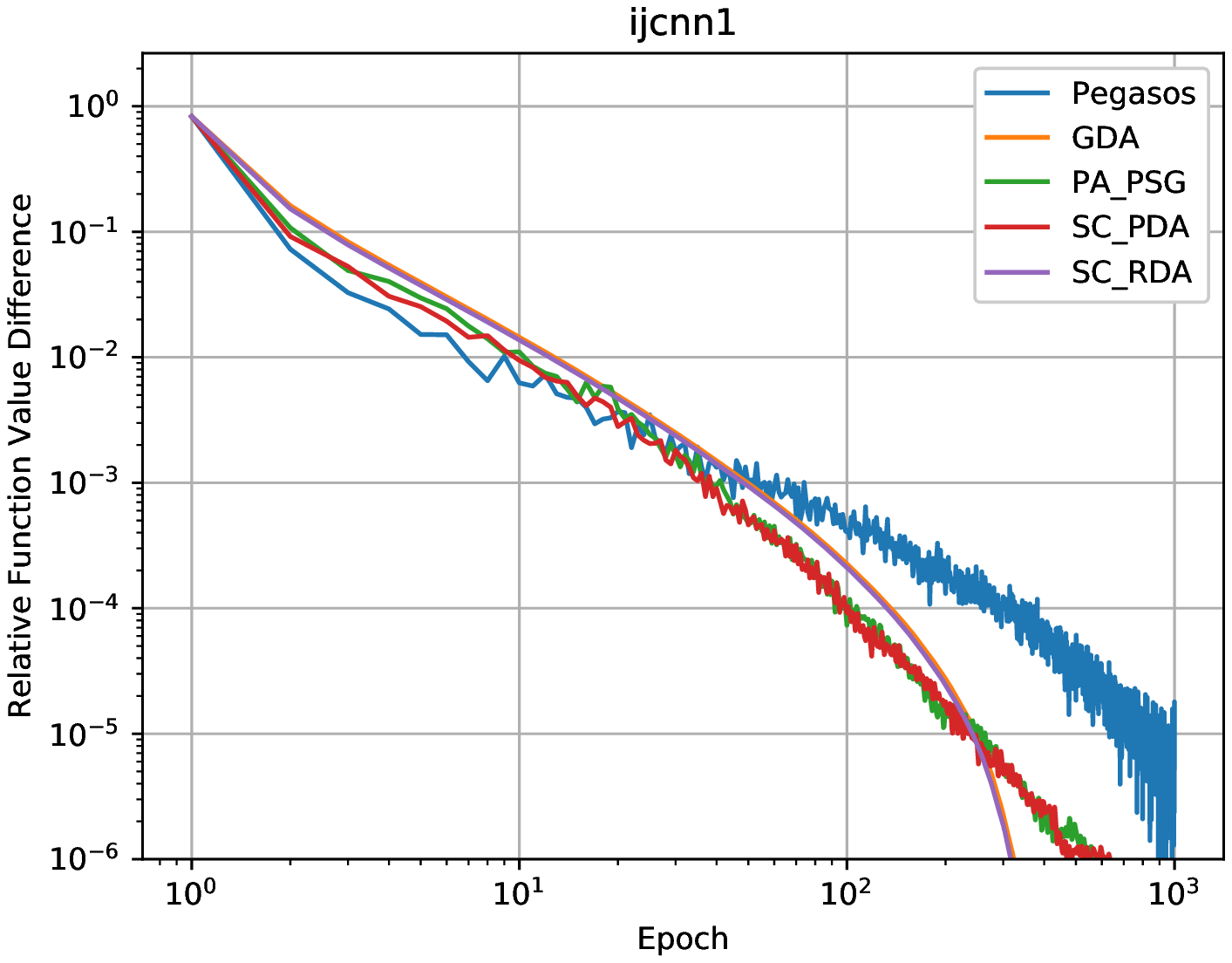}}
    \subfigure[]{
    \includegraphics[width=1.6in]{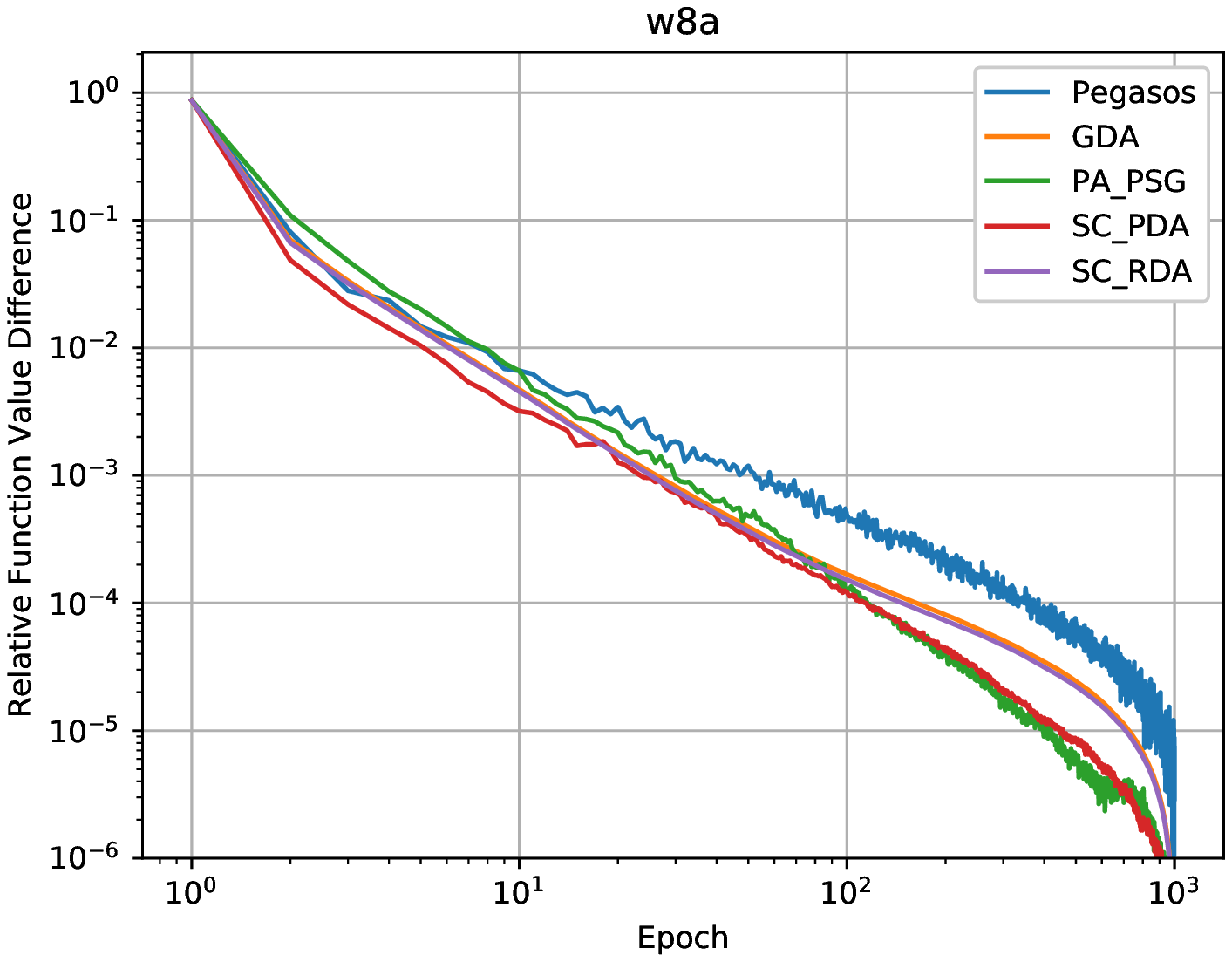}}
    \subfigure[]{
    \includegraphics[width=1.6in]{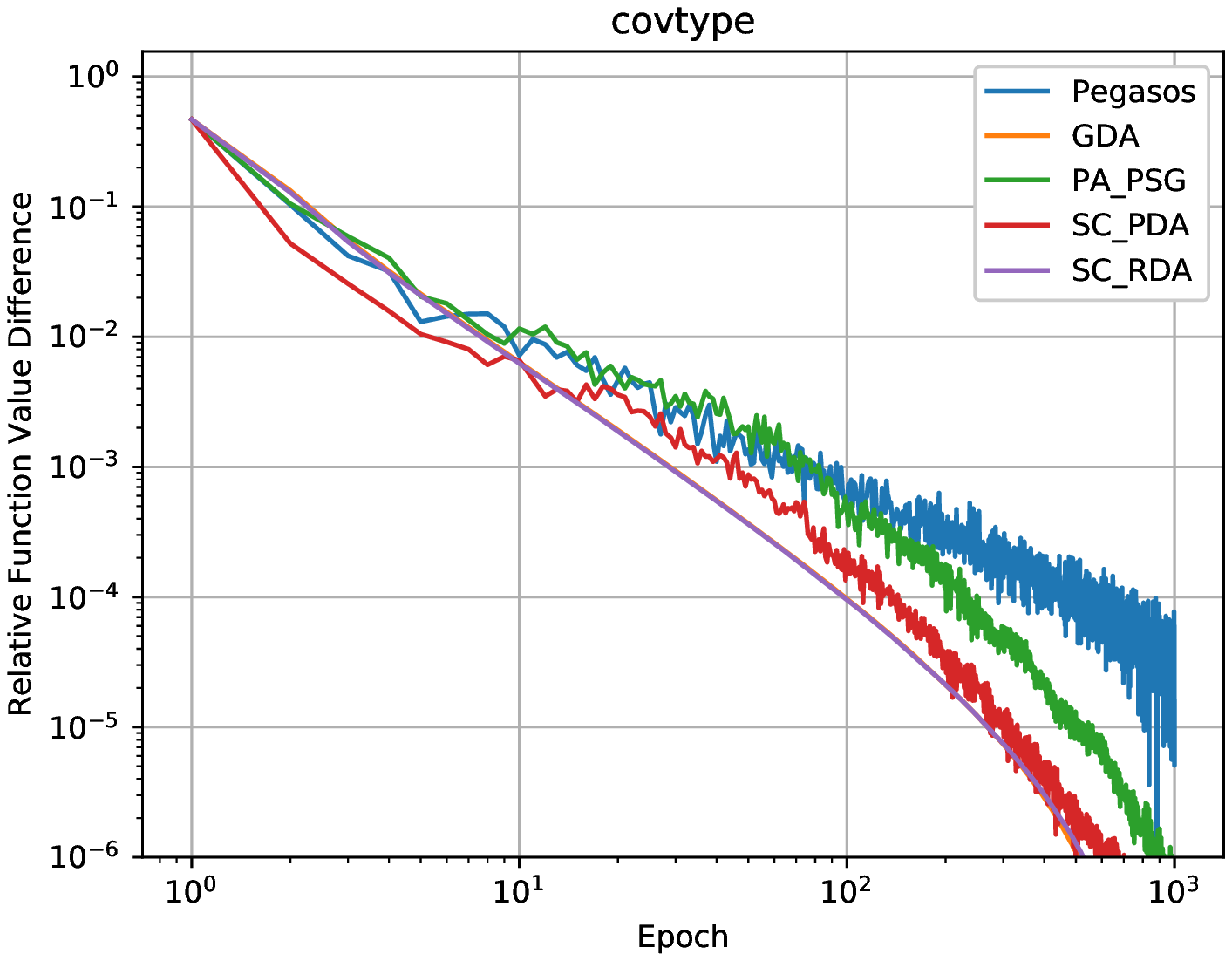}}
    \caption{Convergence on different LibSVM datasets for SVMs}
    \label{fig:convex experiment}
\end{figure*}

\begin{figure*}[ht]
    \centering
    \subfigure[]{
    \includegraphics[width=1.6in]{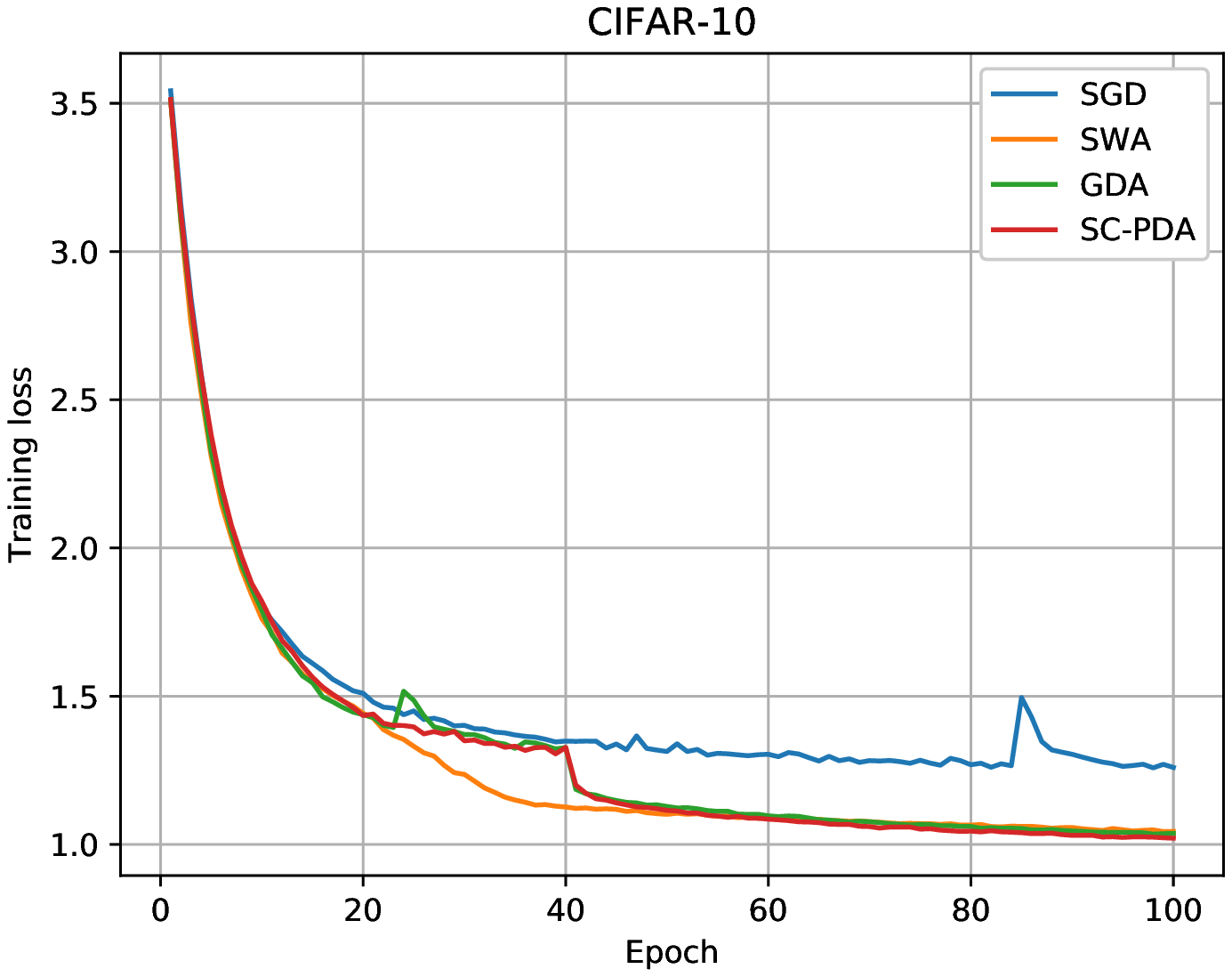}}
    \subfigure[]{
    \includegraphics[width=1.6in]{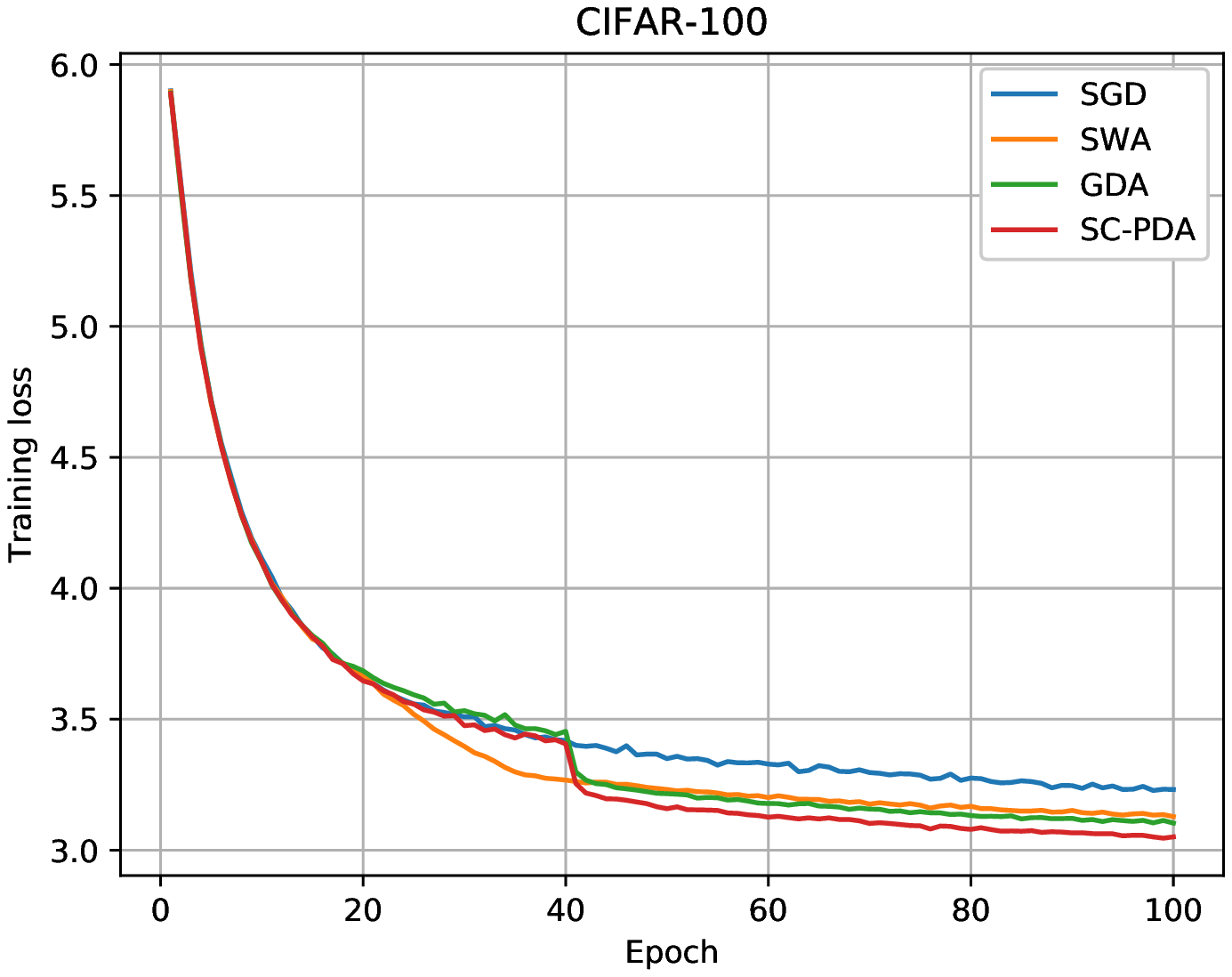}}
    \subfigure[]{
    \includegraphics[width=1.6in]{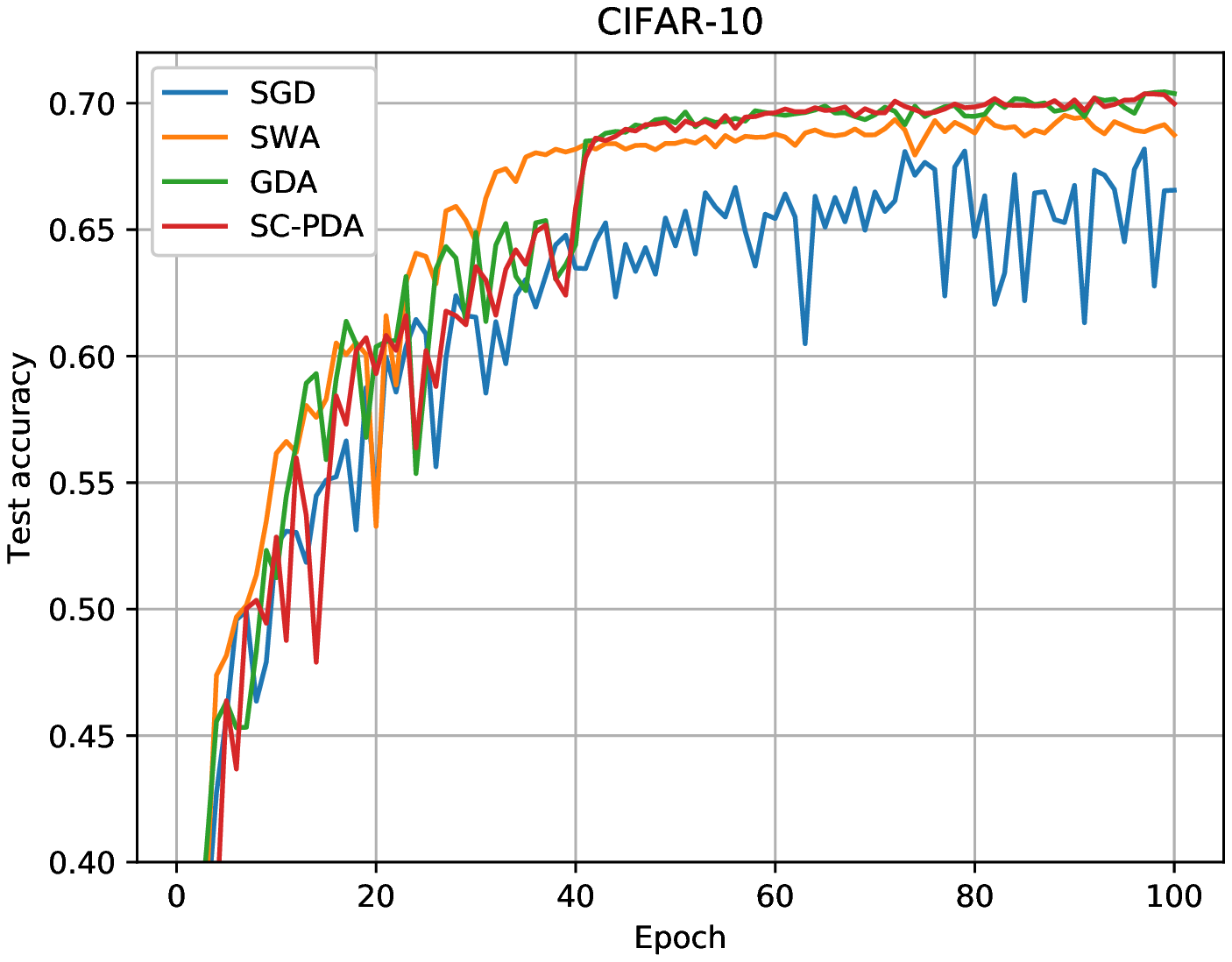}}
    \subfigure[]{
    \includegraphics[width=1.6in]{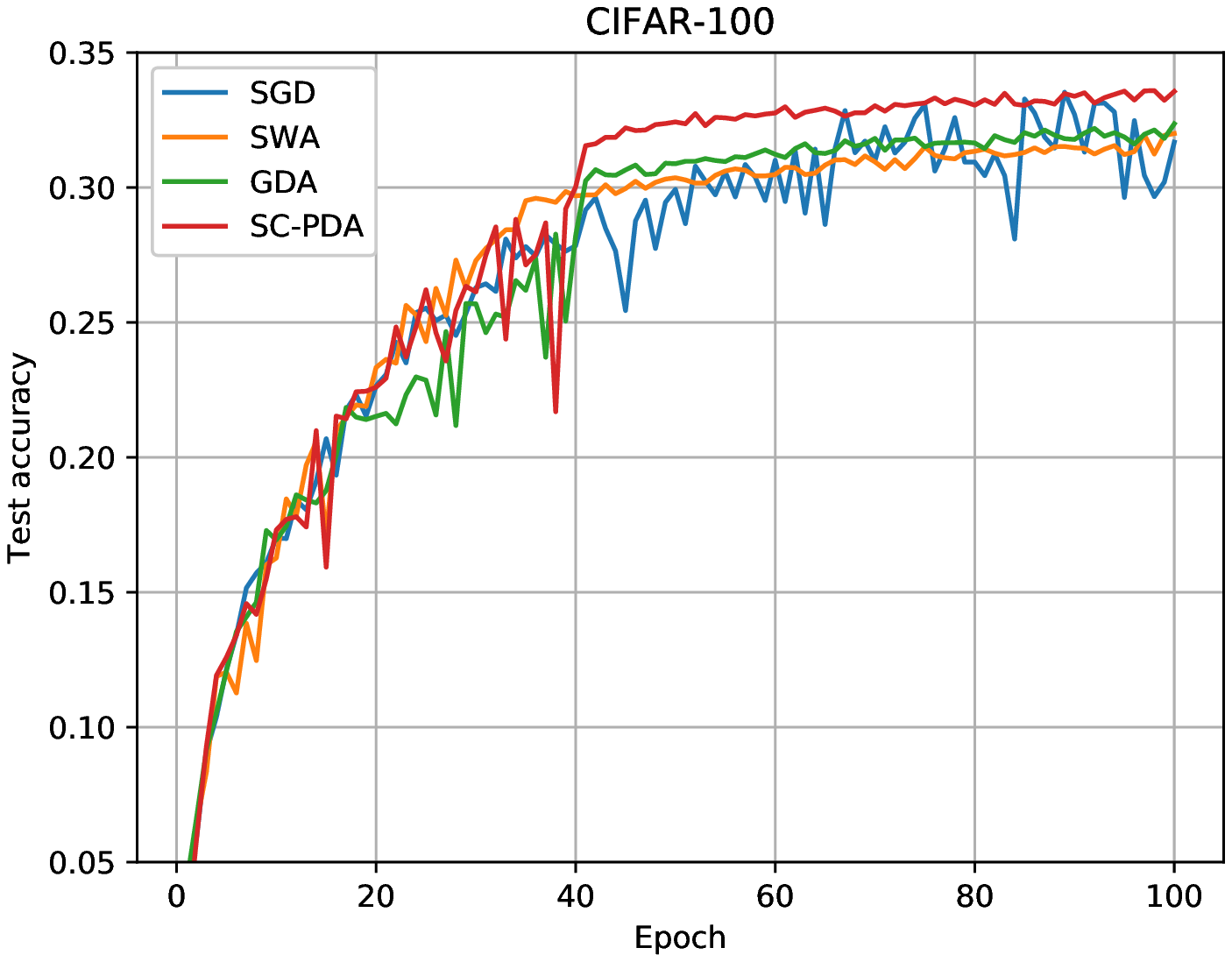}}
    \caption{Training loss and test accuracy on CIFAR-10 and CIFAR-100 datasets for deep learning tasks}
    \label{fig:4cnn}
\end{figure*}

Figure \ref{fig:convex experiment}. exposes how the relative function values $f(\mathbf{w}_t)-f(\mathbf{w_*})$ are changed with respect to the algorithm epoch. As expected, the convergence behavior of our proposed GDA and SC-PDA is very similar to PA-PSG and SC-RDA with the same strong convexity parameter, and lower than Pegasos. Intuitively, the proposed SC-PDA, Pegasos and PA-PSG have almost the same convergence behavior with oscillation, which result from non-averaging scheme. Thus, we derive conclusions that our proposed two algorithms no longer suffer from the additional logarithmic factor and achieve their desired optimal rate.

\subsection{Training Deep Neural Networks}

The second experiment is to show that the proposed algorithms improve the performance of training deep networks. Interestingly, averaging in deep learning has been discussed in a series work about SWA. But they have not observed that the connection between DA and GD with averaging scheme. Moreover, averaging scheme does good for the convergence behavior of the gradient-based methods in terms of individual iterate instead of averaged output. According to the choice of $a_t$ and $\gamma_t$ in the Corollary 2, it should be mentioned that we use the different weighted averaging scheme in GDA and SC-PDA.

Following \cite{izmailov2018averaging, mukkamala2017variants, wang2019sadam}, we conduct experiments on a sever with 2 NVIDIA 2080Ti GPUs. We first design a simple 4-layer CNN architecture that consists two convolutional layers (32 filters of size 3 $\times$ 3), one max-pooling layer (2 $\times$ 2 window and 0.25 dropout) and one fully connected layer (128 hidden units and 0.5 dropout). We also use weight decay with a regularization parameter of 5e-3. The loss function is the cross-entropy. To conduct a fair comparison, the constant learning rate is tuned in \{0.1; 0.01; 0.001; 0.0001\}, and the best results are reported. The training loss and test accuracy are shown in Figure \ref{fig:4cnn}.

Our SC-PDA, GDA and SWA obtain almost the same training loss, lower than SGD (without momentum) on CIFAR-10 and CIFAR-100 datasets\footnote{\url{http://www.cs.toronto.edu/~kriz/cifar.html}}. Moreover, the improved performance also translates into good result on test accuracy. The performance of test accuracy gives our SC-PDA and GDA a slight edge over SGD and SWA. Compared to SGD, averaging schemes always reduce oscillation issues and achieve improvement in better generalization. Although our proposed GDA and SC-PDA are designed for strongly convex functions, it could also lead to practical performance even in some non-convex deep learning tasks.

\section{Conclusion}

In this paper, our original intention is to derive the optimal individual convergence of DA in the strongly convex case. We first propose a general algorithm GDA and we further slightly modify Nesterov's PDA into SC-PDA. We prove that GDA yields the optimal convergence rate in terms of OA, while SC-PDA derives the optimal individual convergence.

Averaging scheme has been frequently employed in modern machine learning community for improving stability and generalization. Unfortunately, there is still a lack of valuable hints how various parameters should be chosen so far. Our proposed GDA and SC-PDA algorithms not only fix the optimal convergence issues but also lead to better performance in deep learning tasks.

\section{Acknowledgements}
This work was supported in part by National Natural Science Foundation of China under Grants (62076252, 61673394, 61976213, 62076251) and in part by Beijing Advanced Discipline Fund.

\bibliography{reference}


\clearpage
\onecolumn

\section{Supplementary Material}
\subsection{Proof of Theorem 1}

According to the strong convexity of $f$ (Definition 1) and 3-point property, 
$$
\begin{aligned}
& \sum_{k=0}^{t} a_{k}\nabla f(\mathbf{w}_k) - \sum_{k=0}^{t} a_{k}\nabla f(\mathbf{w}) \\
& \leq \sum_{k=0}^{t}\langle a_{k}\nabla f(\mathbf{w}_k),\mathbf{w}_k - \mathbf{w}\rangle - \frac{\mu}{2}\sum_{k=0}^{t} a_{k}\|\mathbf{w}_k-\mathbf{w} \|^2 - \frac{\mu}{2}\sum_{k=0}^{t}\gamma_{k}\|\mathbf{w}_k-\mathbf{w} \|^2 + \frac{\mu}{2}\sum_{k=0}^{t}\gamma_{k}\|\mathbf{w}_k-\mathbf{w} \|^2 \\
& \leq \sum_{k=0}^{t}\langle a_{k}\nabla f(\mathbf{w}_k),\mathbf{w}_k - \mathbf{w}_{t+1} \rangle - \frac{\mu}{2}\sum_{k=0}^{t} \gamma_{k}\| \mathbf{w}_k-\mathbf{w}_{t+1} \|^2 - \frac{\mu}{2}\sum_{k=0}^{t} \gamma_{k}\| \mathbf{w}-\mathbf{w}_{t+1} \|^2 + \frac{\mu}{2}\sum_{k=0}^{t}(\gamma_k-a_{k})\|\mathbf{w}_k-\mathbf{w} \|^2.
\end{aligned}
$$
Note that $\sum_{k=0}^{t}\langle a_{k}\nabla f(\mathbf{w}_k),\mathbf{w}_k - \mathbf{w}\rangle-\frac{\mu}{2}\sum_{k=0}^{t} \gamma_{k}\| \mathbf{w}_k-\mathbf{w} \|^2$ is
the objective function of optimization problems in GDA. Then, we consider the following two terms,
$$
\begin{aligned}
& \sum_{k=0}^{t}\langle a_{k}\nabla f(\mathbf{w}_k),\mathbf{w}_k - \mathbf{w}_{t+1} \rangle - \frac{\mu}{2}\sum_{k=0}^{t} \gamma_{k}\| \mathbf{w}_k-\mathbf{w}_{t+1} \|^2  \\
& = \sum_{k=0}^{t-1}\langle a_{k}\nabla f(\mathbf{w}_k),\mathbf{w}_k - \mathbf{w}_{t+1}\rangle - \frac{\mu}{2}\sum_{k=0}^{t-1} \gamma_{k} \| \mathbf{w}_k-\mathbf{w}_{t+1} \|^2 +\langle a_t \nabla f(\mathbf{w}_t),\mathbf{w}_{t} - \mathbf{w}_{t+1}\rangle - \frac{\mu}{2} \gamma_{t} \| \mathbf{w}_t-\mathbf{w}_{t+1} \|^2  \\
& \leq \sum_{k=0}^{t-1}\langle a_{k}\nabla f(\mathbf{w}_k),\mathbf{w}_k - \mathbf{w}_t\rangle - \frac{\mu}{2}\sum_{k=0}^{t-1} \gamma_{k} \| \mathbf{w}_k-\mathbf{w}_{t} \|^2 - \frac{\mu}{2}\sum_{k=0}^{t-1} \gamma_{k} \| \mathbf{w}_t-\mathbf{w}_{t+1} \|^2 \\
& +\langle a_t \nabla f(\mathbf{w}_t),\mathbf{w}_{t} - \mathbf{w}_{t+1}\rangle - \frac{\mu}{2} \gamma_{t} \| \mathbf{w}_t-\mathbf{w}_{t+1} \|^2\\
& \leq \big[\sum_{k=0}^{t-1}\langle a_{k}\nabla f(\mathbf{w}_k),\mathbf{w}_k - \mathbf{w}_t \rangle - \frac{\mu}{2}\sum_{k=0}^{t-1} \gamma_k \| \mathbf{w}_k-\mathbf{w}_{t} \|^2  \big] +\langle a_t \nabla f(\mathbf{w}_t),\mathbf{w}_{t} - \mathbf{w}_{t+1}\rangle - \frac{\mu}{2}\sum_{k=0}^{t} \gamma_{k} \| \mathbf{w}_t-\mathbf{w}_{t+1} \|^2.
\end{aligned}
$$
Further, we can combine last two items with Fenchel-Young inequality, such that
$$
\begin{aligned}
& a_{t}\langle \nabla f(\mathbf{w}_{t}),\mathbf{w}_{t} - \mathbf{w}_{t+1} \rangle - \frac{\mu}{2}\sum_{k=0}^{t} \gamma_{k} \| \mathbf{w}_t-\mathbf{w}_{t+1} \|^2 \leq \frac{a_t^2}{2 \mu \Gamma_t} \|\nabla f(\mathbf{w}_t)\|^2 ,
\end{aligned}
$$
where $\Gamma_t = \sum_{k=0}^{t} \gamma_k$. Applying recursively this inequality, we can get
$$
\begin{aligned}
\frac{1}{A_t} & \sum_{k=0}^{t}a_k f(\mathbf{w}_k)-f(\mathbf{w}_*) \leq  \frac{1}{A_t}\sum_{k=0}^{t}\frac{a_{k}^{2}}{2\mu \Gamma_k}\|\nabla f(\mathbf{w}_k)\|^2 + \frac{\mu}{2A_t}\sum_{k=0}^{t}(\gamma_k-a_{k})\|\mathbf{w}_k-\mathbf{w} \|^2
\end{aligned}
$$
Thus, Theorem 1 is proved.

\subsection{Proof of Lemma 3}

According to the step of PA, we note that $$
a_t(\mathbf{w}_t - \mathbf{w}_{t-1}^+) = A_{t-1}\left(\mathbf{w}_{t-1}-\mathbf{w}_t\right),
$$  
then let's multiply both sides by $\nabla f(\mathbf{w}_t) $, and the left side of the above equality becomes
$$
\begin{aligned}
a_t \langle \nabla f(\mathbf{w}_t), \mathbf{w}_t-\mathbf{w}_{t}^+ \rangle 
= a_t \langle \nabla f(\mathbf{w}_t), \mathbf{w}_t-\mathbf{w}_{t-1}^+ \rangle+ a_t \langle \nabla f(\mathbf{w}_t), \mathbf{w}_{t-1}^+-\mathbf{w}_{t}^+ \rangle,
\end{aligned}
$$
Next, let's consider the first inner product term,
$$
\begin{aligned}
& a_t \langle \nabla f(\mathbf{w}_t), \mathbf{w}_t-\mathbf{w}_{t-1}^+ \rangle
\\ & = A_{t-1} \langle \nabla f(\mathbf{w}_t), \mathbf{w}_{t-1}-\mathbf{w}_{t} \rangle
\\ & \leq A_{t-1}f(\mathbf{w}_{t-1})-A_{t-1}f(\mathbf{w}_{t}) - \dfrac{\mu}{2} A_{t-1} \| \mathbf{w}_{t-1} - \mathbf{w}_t\|^2.
\end{aligned}
$$
Finally, we have
$$
\begin{aligned}
 &  a_t \langle \nabla f(\mathbf{w}_t),\mathbf{w}_t - \mathbf{w}_t^+ \rangle
\\  & =  a_t \langle \nabla f(\mathbf{w}_t),\mathbf{w}_t - \mathbf{w}_{t-1}^+ \rangle +  a_t \langle \nabla f(\mathbf{w}_t),\mathbf{w}_{t-1}^+ - \mathbf{w}_{t}^+\rangle
\\ & = A_{t-1} \langle \nabla f(\mathbf{w}_t),\mathbf{w}_t - \mathbf{w}_{t-1} \rangle + a_t \langle \nabla f(\mathbf{w}_t),\mathbf{w}_{t-1}^+ - \mathbf{w}_{t}^+\rangle
\\ & \leq A_{t-1} [f(\mathbf{w}_{t-1}) - f(\mathbf{w}_{t})] + a_t \langle \nabla f(\mathbf{w}_t),\mathbf{w}_{t-1}^+ - \mathbf{w}_{t}^+\rangle.
 \end{aligned}
$$
Lemma 3 is proved.

\subsection{Proof of Theorem 2}

Similar to the proof of Theorem 1, 
$$
\begin{aligned}
& \sum_{k=1}^{t} a_{k}\nabla f(\mathbf{w}_k) - \sum_{k=1}^{t} a_{k}\nabla f(\mathbf{w})
\\ & \leq \sum_{k=1}^{t}\langle a_{k}\nabla f(\mathbf{w}_k),\mathbf{w}_k - \mathbf{w}\rangle - \frac{\mu}{2}\sum_{k=1}^{t} a_{k}\|\mathbf{w}_k-\mathbf{w} \|^2 - \frac{\mu}{2}\sum_{k=1}^{t}\gamma_{k}\|\mathbf{w}_k-\mathbf{w} \|^2 + \frac{\mu}{2}\sum_{k=1}^{t}\gamma_{k}\|\mathbf{w}_k-\mathbf{w} \|^2
\\ & \leq \sum_{k=1}^{t}\langle a_{k}\nabla f(\mathbf{w}_k),\mathbf{w}_k - \mathbf{w}_t^+ \rangle - \frac{\mu}{2}\sum_{k=1}^{t} \gamma_{k}\| \mathbf{w}_k-\mathbf{w}_t^+ \|^2  - \frac{\mu}{2} \sum_{k=1}^{t} \gamma_{k}\| \mathbf{w}-\mathbf{w}_t^+ \|^2 + \frac{\mu}{2} \sum_{k=1}^{t} (\gamma_{k}-a_k)\| \mathbf{w}-\mathbf{w}_t^+ \|^2.
\end{aligned}
$$

Next, we consider the first two terms,
$$
\begin{aligned}
& \sum_{k=0}^{t}\langle a_{k}\nabla f(\mathbf{w}_k),\mathbf{w}_k - \mathbf{w}_t^+ \rangle  - \frac{\mu}{2} \sum_{k=0}^{t} \gamma_{k}\| \mathbf{w}_k-\mathbf{w}_t^+ \|^2 
\\ & = \sum_{k=1}^{t-1}\langle a_{k}\nabla f(\mathbf{w}_k),\mathbf{w}_k - \mathbf{w}_t^+ \rangle  - \frac{\mu}{2}\sum_{k=0}^{t-1} \gamma_{k}\| \mathbf{w}_k-\mathbf{w}_t^+ \|^2  -\frac{\mu}{2} \gamma_t \| \mathbf{w}_t -\mathbf{w}_t^+ \|^2 + \langle a_{t}\nabla f(\mathbf{w}_t),\mathbf{w}_t - \mathbf{w}_t^+ \rangle
\\ & \leq \sum_{k=1}^{t-1}\langle a_{k}\nabla f(\mathbf{w}_k),\mathbf{w}_k - \mathbf{w}_{t-1}^+ \rangle  - \frac{\mu}{2}\sum_{k=0}^{t-1} \gamma_{k}\| \mathbf{w}_k-\mathbf{w}_{t-1}^+ \|^2 - \frac{\mu}{2}\sum_{k=0}^{t-1} \gamma_{k}\| \mathbf{w}_t^+ - \mathbf{w}_{t-1}^+ \|^2 \\
& + \langle a_{t} \nabla f(\mathbf{w}_t),\mathbf{w}_t -\mathbf{w}_t^+ \rangle -\frac{\mu}{2} \gamma_t \| \mathbf{w}_t -\mathbf{w}_t^+ \|^2
\end{aligned}
$$

Then, according to Lemma 3, we have
 $$
\begin{aligned}
& \leq \sum_{k=1}^{t-1}\langle a_{k}\nabla f(\mathbf{w}_k),\mathbf{w}_k - \mathbf{w}_{t-1}^+ \rangle - \frac{\mu}{2}\sum_{k=0}^{t-1} \gamma_{k}\| \mathbf{w}_k-\mathbf{w}_{t-1}^+ \|^2 - \frac{\mu}{2} \sum_{k=0}^{t-1} \gamma_{k}\| \mathbf{w}_t^+ - \mathbf{w}_{t-1}^+ \|^2 \\
& - \frac{\mu}{2} \gamma_{t}\| \mathbf{w}_t^+ - \mathbf{w}_{t-1}^+ \|^2 + \langle a_{t}\nabla f(\mathbf{w}_t),\mathbf{w}_{t-1}^+ - \mathbf{w}_t^+ \rangle + A_{t-1}\big[f(\mathbf{w}_{t-1})-f(\mathbf{w}_t)\big]  \\
& \leq A_{t-1}\big[f(\mathbf{w}_{t-1})-f(\mathbf{w}_t)\big] +\big[ \sum_{k=0}^{t-1}\langle a_{k}\nabla f(\mathbf{w}_k),\mathbf{w}_k - \mathbf{w}_{t-1}^+ \rangle - \frac{\mu}{2}\sum_{k=0}^{t-1} \gamma_{k}\| \mathbf{w}_k-\mathbf{w}_{t-1}^+ \|^2 \big] \\
& + \big[ a_{t}\langle \nabla f(\mathbf{w}_{t}),\mathbf{w}_{t-1}^ + - \mathbf{w}_t^+ \rangle
 - \frac{\mu}{2} \Gamma_{t} \| \mathbf{w}_t^ + - \mathbf{w}_{t-1}^+\|^2\big]\\
& \leq A_{t-1}\big[f(\mathbf{w}_{t-1})-f(\mathbf{w}_t)\big] +\big[ \sum_{k=0}^{t-1}\langle a_{k}\nabla f(\mathbf{w}_k),\mathbf{w}_k - \mathbf{w}_{t-1}^+ \rangle - \frac{\mu}{2}\sum_{k=0}^{t-1} \gamma_{k}\| \mathbf{w}_k-\mathbf{w}_{t-1}^+ \|^2 \big] \\
& + \frac{a_t^2}{2\mu \Gamma_t} \|\nabla f(\mathbf{w}_t)\|^2
\end{aligned}
$$
Then applying recursively this inequality for $A_t [f(\mathbf{w}_t)-f(\mathbf{w})]$, Theorem 2 is proved.

\subsection{Proof of Corollary 3}

Note that 
$$
\bar{\mathbf{w}}_t = \dfrac{1}{t}\sum_{i=0}^{n}\mathbf{w}_i
$$
is called standard averaging scheme in mathematics. According to Corollary 2 (1) and using the property of Harmonic Serie, we can get 
$$f(\bar{\mathbf{w}}_t)-f(\mathbf{w}_*)\leq O\left(\frac{\log t}{t}\right).$$
Thus, Corollary 2 is proved.

\subsection{Proof of Theorem 3}

We just easily replace the gradient operation $\nabla f(\mathbf{w}_t) $ with $\hat{\mathbf{g}}_t$ in the stochastic setting, because of theoretical guarantees \cite{rakhlin2011making}. Therefore, we can have
$$
\begin{aligned}
& \mathbb{E}[f(\mathbf{w}_t) - f(\mathbf{w})]
 \leq \frac{1}{A_t}\sum_{k=0}^{t}\frac{a_{k}^{2}}{2\mu \Gamma_k}\|\hat{\mathbf{g}}_t\|^2 + \frac{\mu}{2A_t}\sum_{k=0}^{t}(\gamma_k-a_{k})\|\mathbf{w}_k-\mathbf{w} \|^2.
\end{aligned}
$$
Then, Let $a_t=\gamma_t=t$, it holds that
$$ \mathbb{E}[f(\mathbf{w}_t)-f(\mathbf{w}) ]\leq O\left(\frac{1}{t}\right).$$
Theorem 3 can be proved.

\begin{figure}[ht]
    \centering
    \subfigure[]{
    \includegraphics[width=0.4\textwidth]{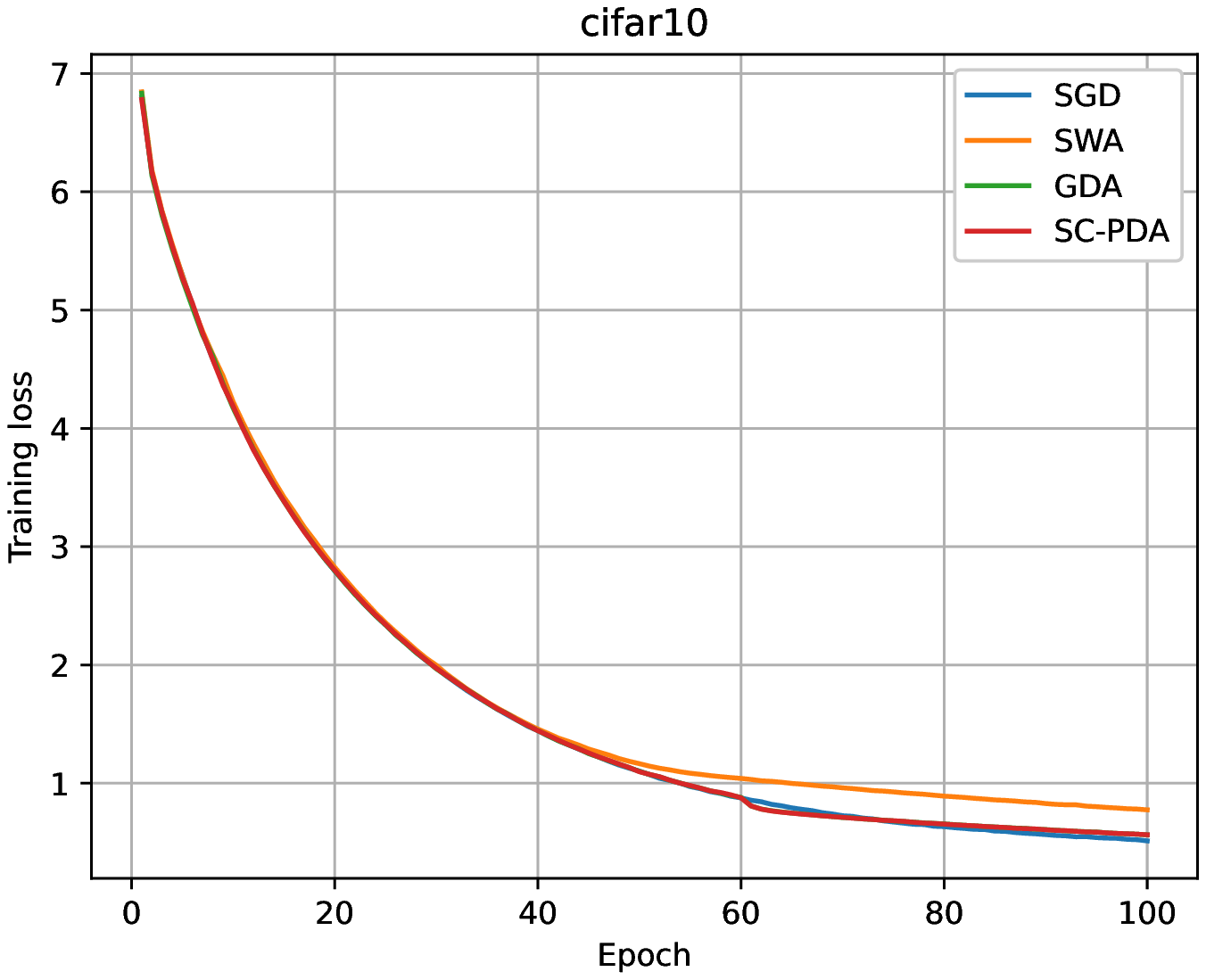}}
    \subfigure[]{
    \includegraphics[width=0.4\textwidth]{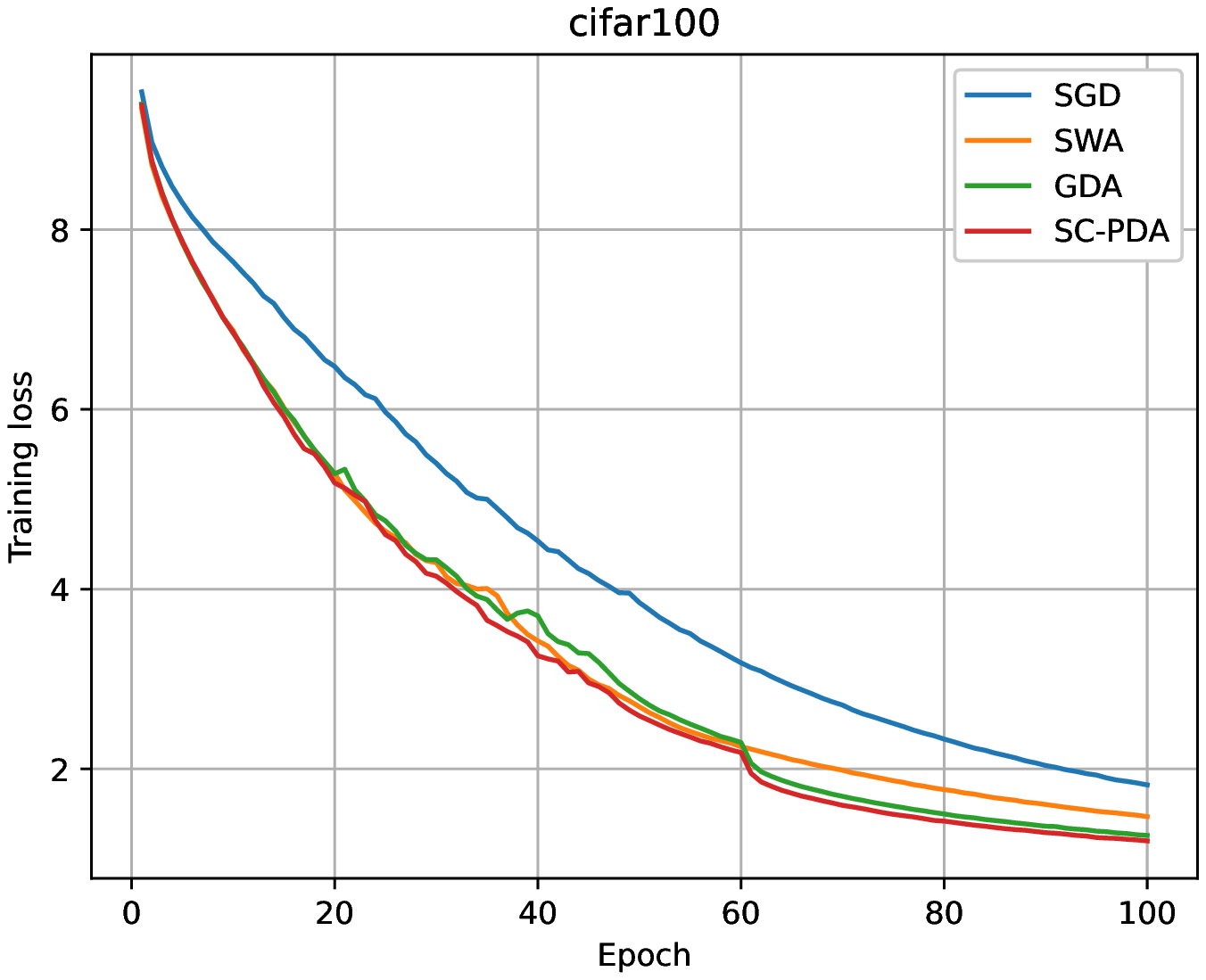}}\\
    \subfigure[]{
    \includegraphics[width=0.4\textwidth]{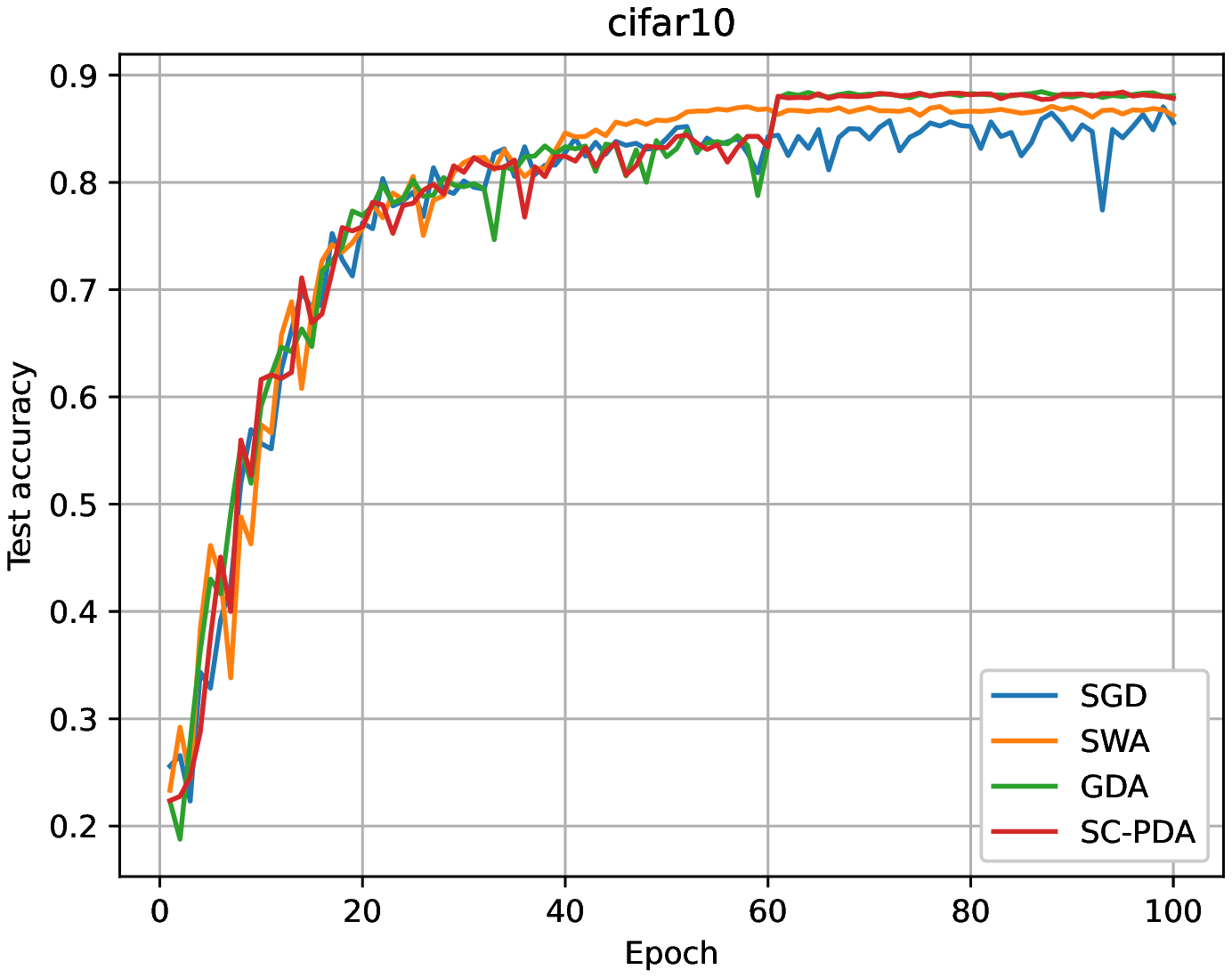}}
    \subfigure[]{
    \includegraphics[width=0.4\textwidth]{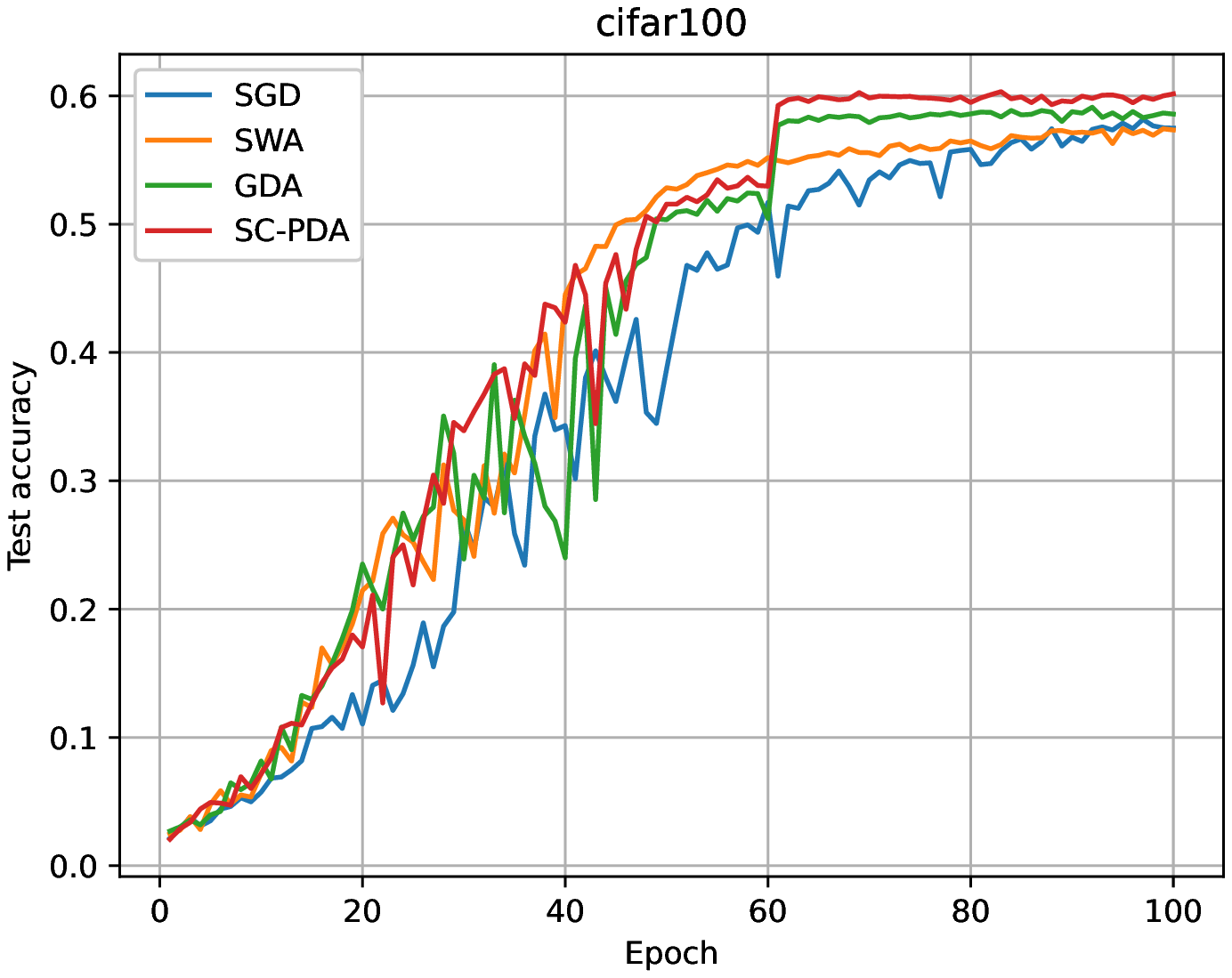}}
    \caption{Training loss and test accuracy on CIFAR-10 and CIFAR-100 datasets for VGG-16}
    \label{fig:vgg}
\end{figure} 

\subsection{Additional Experiment Results}

To further emphasize the performance of our proposed GDA and SC-PDA, we also conduct experiments on standard VGG-16. For fair comparison, the constant learning rate is tuned in \{0.1; 0.05; 0.01; 0.005; 0.001\} and the best results are reported. We also repeat each experiment five times and take their average results. As can be seen in figure \ref{fig:vgg}, the performance of our proposed GDA and SC-PDA is better than that of SWA and SGD. We obtain the conclusion that algorithms coupled with averaging schemes in deep learning always reduce oscillation issues and achieve improvement in test accuracy.

\end{document}